\documentclass[11pt]{article}

\usepackage[final]{ACL}

\usepackage{times}
\usepackage{latexsym}
\usepackage{booktabs}
\usepackage[T1]{fontenc}

\usepackage[utf8]{inputenc}

\usepackage{microtype}

\usepackage{inconsolata}

\usepackage{graphicx}
\usepackage{amsmath}
\usepackage{amssymb}
\usepackage{algorithm}
\usepackage{algorithmic}
\usepackage{tcolorbox}
\tcbuselibrary{listings,skins,breakable}
\usepackage{float}

\usepackage{xcolor}

\usepackage{xspace}  
\newcommand{\agent}{\textsc{CAFES}\xspace}
\usepackage{enumitem}

\usepackage{pifont}

\usepackage{graphicx}   
\usepackage{booktabs}   
\usepackage{multirow}   
\usepackage{arydshln}   
\usepackage{array}
\usepackage{tabularx}

\definecolor{green2}{RGB}{90, 130, 60}
\definecolor{gray2}{RGB}{169, 169, 169}
\definecolor{red2}{RGB}{178, 34, 34}

\newcommand{\up}[1]{%
  \mbox{\textcolor{green2}{\small ↑#1}}%
}
\newcommand{\down}[1]{%
  \mbox{\textcolor{red2}{\small ↓#1}}%
}

\usepackage{bm}
\title{\raisebox{-0.11cm}{\includegraphics[width=0.15\textwidth]{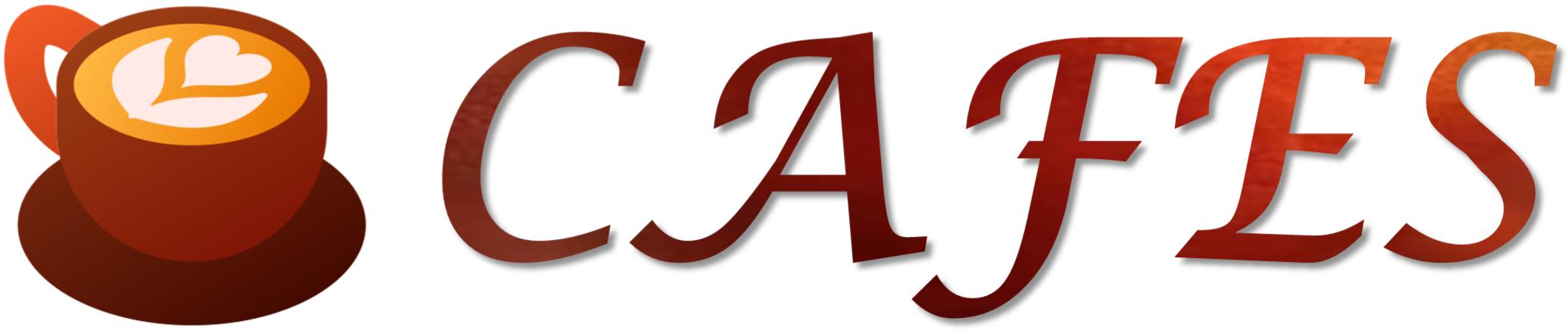}}: A Collaborative Multi-Agent Framework \\for Multi-Granular Multimodal Essay Scoring}

\author{Jiamin Su\textsuperscript{\rm 1,\rm 2}, Yibo Yan\textsuperscript{\rm 1,\rm 3}, Zhuoran Gao\textsuperscript{\rm 1}, \textbf{Han Zhang}\textsuperscript{\rm 1},
\textbf{Xiang Liu}\textsuperscript{\rm 1,\rm 3},
\textbf{Xuming Hu}\textsuperscript{\rm 1,\rm 2,\rm 3,}\footnotemark[2]\\
\fontsize{9.0pt}{\baselineskip}\selectfont \textsuperscript{\rm 1}The Hong Kong University of Science and Technology (Guangzhou)\\
\fontsize{9.0pt}{\baselineskip}\selectfont \textsuperscript{\rm 2}Beijing Future Brain Education Technology Co., Ltd.\\
\fontsize{9.0pt}{\baselineskip}\selectfont \textsuperscript{\rm 3}The Hong Kong University of Science and Technology\\
\fontsize{9.0pt}{\baselineskip}\selectfont \texttt{jsu360@connect.hkust-gz.edu.cn},
\texttt{xuminghu@hkust-gz.edu.cn}
}

\begin{document}
\maketitle
\begin{abstract}
Automated Essay Scoring (AES) is crucial for modern education, particularly with the increasing prevalence of multimodal assessments.
However, traditional AES methods struggle with \textit{evaluation generalizability and multimodal perception}, while even recent Multimodal Large Language Model (MLLM)-based approaches can produce \textit{hallucinated justifications and scores misaligned with human judgment}.
To address the limitations, we introduce \textbf{\agent}, the first \underline{c}ollaborative multi-\underline{a}gent \underline{f}ramework specifically designed for A\underline{E}\underline{S}.
It orchestrates three specialized agents: an \textit{Initial Scorer} for rapid, trait-specific evaluations; a \textit{Feedback Pool Manager} to aggregate detailed, evidence-grounded strengths; and a \textit{Reflective Scorer} that iteratively refines scores based on this feedback to enhance human alignment.
Extensive experiments, using state-of-the-art MLLMs, achieve an average relative improvement of 21\% in Quadratic Weighted Kappa (QWK) against ground truth, especially for grammatical and lexical diversity. Our proposed \agent framework paves the way for an intelligent multimodal AES system. The code will be available upon acceptance.
\end{abstract}

\section{Introduction}
Automated Essay Scoring (AES) plays a crucial role in educational assessment today, offering efficient, fair, and scalable evaluation of student writing tasks \cite{ramesh2022automated,li2024applying,wu2024unveiling,xia2024empirical}. AES systems benefit both students by highlighting areas for improvement and educators by reducing manual grading workloads. As contemporary assessments increasingly emphasize students' abilities to integrate information from both text and images, multimodal writing tasks have become a key focus. Therefore, there is \textit{a growing need for AES systems capable of precise, detailed, context-aware evaluations that effectively handle multimodal inputs} \cite{ye2025position,su2025essayjudge,li2024bringing}.

\begin{figure}[tb!]
  \centering
  \includegraphics[width=0.48\textwidth]{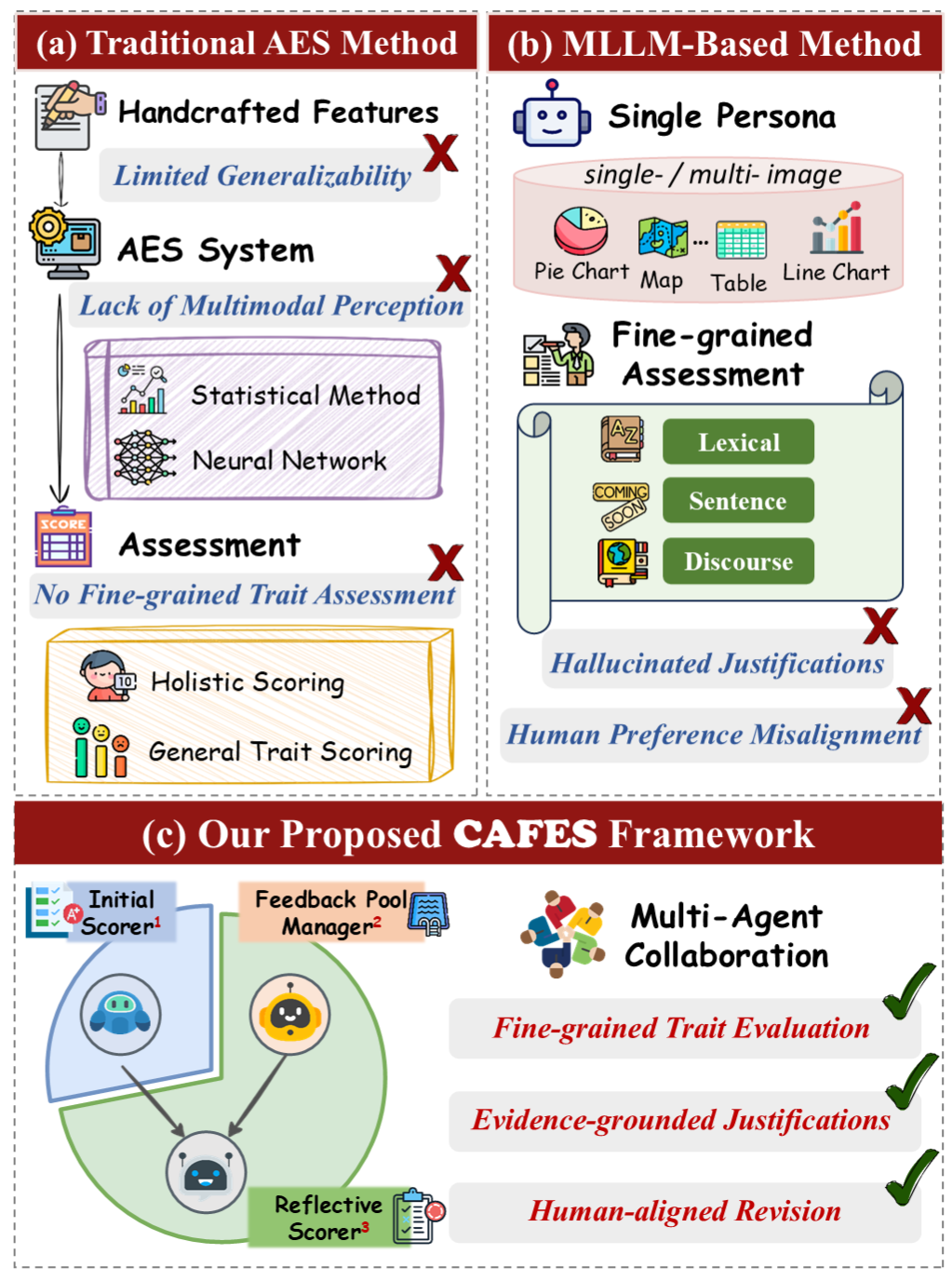}
  \caption{Comparisons among the traditional AES method (a), MLLM-based method (b), and our proposed multi-agent \agent framework (c) on AES task.}
  \label{fig:introduction}
  \vspace{-4mm}
\end{figure}

\textbf{Traditional AES methods} suffer from several critical limitations, as shown in Figure \ref{fig:introduction} (a). \ding{182} They rely heavily on handcrafted features like word frequency and essay length, limiting their generalizability across diverse topics \cite{yang2024unveiling,jansen2024individualizing,uto2020neural}. \ding{183} They lack multimodal perception, making them unsuitable for multimodal inputs. \ding{184} They struggle to assess fine-grained traits, such as coherence and organizational structure \cite{lim2021comprehensive,uto2021review,wang2022bert}. Recently, Multimodal Large Language Models (MLLMs) have been applied to AES, yet \textbf{MLLM-based methods} still introduce new challenges like \ding{185} hallucinated justifications and \ding{186} scoring misaligned with human preference \cite{su2025essayjudge}, as shown in Figure \ref{fig:introduction} (b).

However, the emergence of multimodal multi-agent systems offers a promising solution to these challenges \cite{wang2025comprehensive,chu2025llm}. Specifically, \textbf{multi-agent frameworks} have the following advantages, as illustrated in Figure \ref{fig:introduction} (c): \ding{52} They enable fine-grained trait evaluation, providing detailed feedback across various writing traits. \ding{52} They can generate evidence-grounded justifications and engage in cross-agent collaboration, effectively mitigating hallucinations introduced by a single MLLM. \ding{52} The reflective mechanism of multi-agent systems ensures human-aligned revisions, dynamically adjusting scores to better align with human preferences.

Therefore, we propose \textbf{\agent, the first-ever \underline{c}ollaborative multi‐\underline{a}gent \underline{f}ramework designed specifically for A\underline{E}\underline{S}}. In particular, \agent decomposes the scoring process into three specialized modules: an \textit{initial scoring agent} that provides fast trait‐specific scores; a \textit{feedback pool agent} that aggregates detailed strengths across writing traits; and a \textit{reflective scoring agent} that iteratively updates scores based on the feedback pool.

Our contributions can be summarized as follows:
\begin{itemize}[leftmargin=*]
    \item We introduce \agent, \textbf{the first multi-agent framework} for AES tasks, integrating three specialized agents including Initial scorer, Feedback Pool Manager, and Reflective Scorer to enable collaborative multi-granular essay scoring.
    \item We demonstrate \textbf{the essential impact of the Feedback Pool Manager, Reflective Scorer, and teacher-student MLLM collaboration mechanism} through ablation studies and case analyses.
    \item We evaluate \agent framework with \textbf{state-of-the-art MLLMs} as student models, GPT-4o as the default teacher model, achieving an average improvement of 21\% in Quadratic Weighted Kappa (QWK) against ground truth scores, especially for  grammatical and lexical diversity traits.
\end{itemize}
By addressing the gaps in the existing AES approaches, \agent pave the way for reliable, nuanced, and context-sensitive multi-agent AES systems driven by MLLMs in the AGI era.

\section{Related Work}

\subsection{AES Datasets}
Existing AES datasets have been widely used to support research on writing assessment (more details are shown in the Appendix \ref{sec:dataset}). In terms of modality, these datasets can be categorized into text-only and multimodal datasets.

Among \textbf{text-only datasets}, $\text{ASAP}_{\text{AES}}$ \cite{cozma2018ASAP} is widely used due to its large scale. Its extended version, ASAP++, adds trait-level annotations, but merges key content traits into a single “CONTENT” label \cite{mathias2018asap++}. Both of them only have few topics.
The CLC-FCE dataset provides detailed annotations of grammatical errors \cite{yannakoudakis2011CLCFCE}. The TOEFL11 dataset uses only coarse-grained proficiency labels (low / medium / high) \cite{lee2024TOEFL11}.
The ICLE \cite{granger2009icle} and ICLE++ \cite{li2024icle++} datasets offer more detailed and multi-granular annotations. Nevertheless, their topic coverage is highly limited. Similarly, the AAE corpus focuses solely on argumentative structure \cite{stab2014AAE}. The CREE corpus is designed to evaluate sentence understanding and error types \cite{bailey2008CREE}.
In summary, existing text-only AES datasets face two key limitations: (1) limited topic diversity, and (2) a lack of fine-grained trait-level annotations \cite{Ke2019survey,li2024recent,li2024reflection}. 

EssayJudge is the only publicly available \textbf{multimodal AES dataset} \cite{su2025essayjudge}, with ten fine-grained trait annotation. Lexical-level traits include \textit{lexical accuracy} (LA) and \textit{lexical diversity} (LD). Sentence-level traits include \textit{coherence} (CH), \textit{grammatical accuracy} (GA), \textit{grammatical diversity} (GD), and \textit{punctuation accuracy} (PA). Discourse-level traits include \textit{argument clarity} (AC), \textit{justifying persuasiveness} (JP), \textit{organizational structure} (OS), and \textit{essay length} (EL).

\begin{figure*}[htb!]
  \centering    \includegraphics[width=1\textwidth]{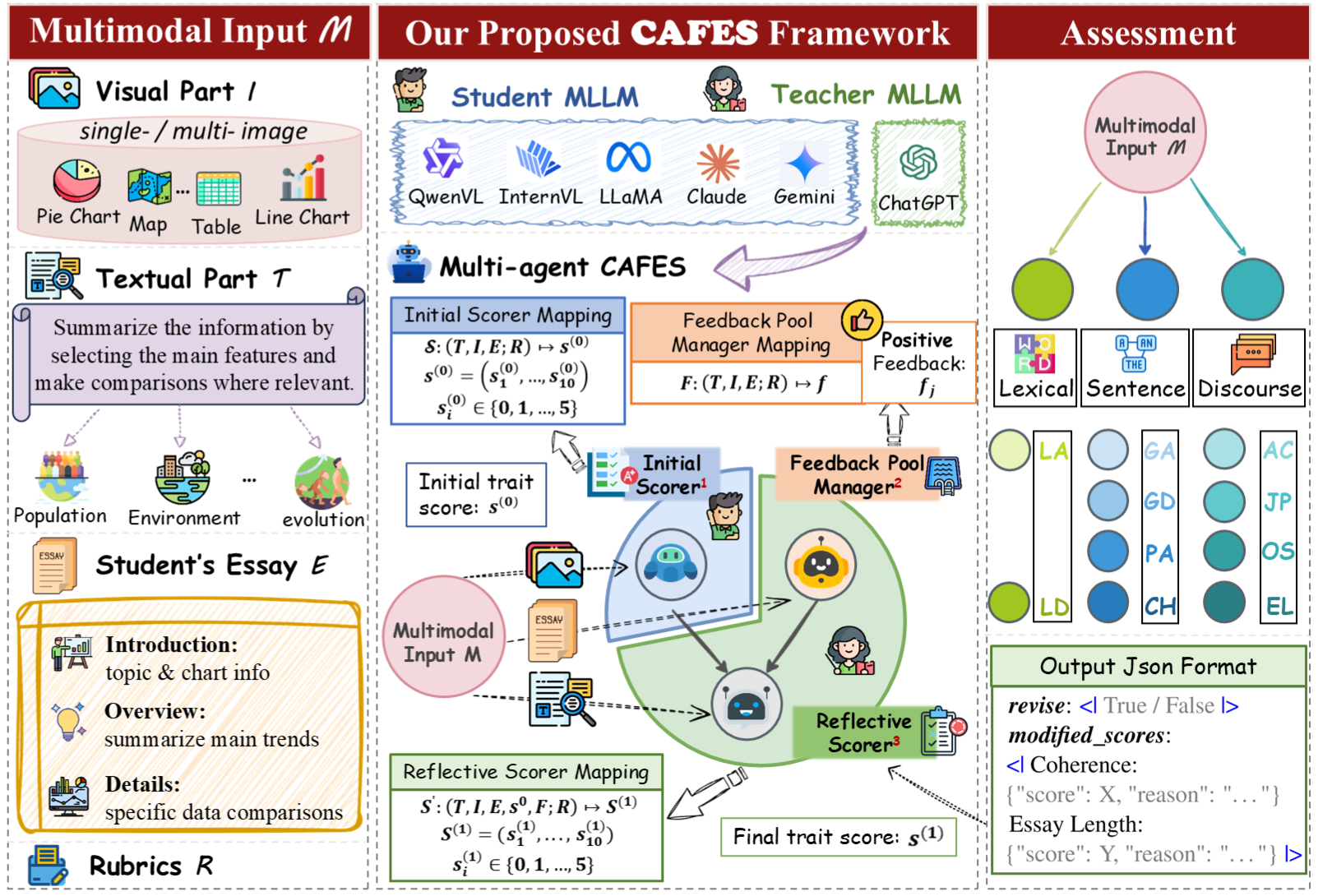}
  \caption{The framework of our proposed \agent.
The system follows a three-stage process: \ding{182} Initial scoring via the student MLLM; \ding{183} Feedback generation for each trait via the teacher MLLM; and \ding{184} Final reflective scoring with justification-based revision via the teacher MLLM.}
  \label{fig:roadmap}
  \vspace{-2mm}
\end{figure*}

\subsection{AES Systems}
AES systems are typically classified into three types: heuristic, machine learning, and deep learning approaches \cite{li2024reflection,kamalov2025evolution,atkinson2025llm,xu2025explainable,song2025unified}.
\textbf{Heuristic methods} assign overall scores by combining rule-based trait scores such as organization, coherence, and grammar. For instance, the organization can be assessed using templates like the three-paragraph essay format \cite{Attali2006erator}.
\textbf{Machine learning methods} (\textit{e.g.,} Logistic Regression, SVMs) rely on handcrafted features \cite{chen2013lexical, yannakoudakis2012lenthbased}, including length-based attributes \cite{Sowmya2016linguistic,yannakoudakis2012lenthbased}. Thus, their cross-topic generalization is limited.
\textbf{Deep learning methods}, especially Transformer models like BERT \cite{wang2022bert}, improve AES by learning semantic features directly from text \cite{jiang2023improving, cao2020domain}, enabling multi-trait and cross-topic scoring.
\textbf{LLM-based methods} further advance AES \cite{MIZUMOTO2023llmbased,choi2025anchor,cai2025rank}. They support zero-shot scoring using rubrics alone \cite{lee2024TOEFL11}, or few-shot settings with minimal examples \cite{mansour2024fewshoting, xiao2024fewshoting}, offering better performance and adaptability in low-resource scenarios.

\subsection{Multi-Agent Collaboration}
Recent studies suggest that multi-agent collaboration, by organizing and coordinating multiple LLMs, enables more effective handling of complex tasks \cite{tran2025multiagentcollaborationmechanismssurvey,yan2025mathagent}. Systems like CAMEL \cite{li2023camelcommunicativeagentsmind} and AutoGen \cite{wu2023autogenenablingnextgenllm} assign roles such as planner, coder, and critic, allowing agents to interact through multi-turn dialogue and perform better in reasoning, generation, and self-revision \cite{liang2024encouraging}.
This approach offers key benefits: improved task decomposition and control through role division, reduced bias via mutual verification, and enhanced adaptability and modularity. It is increasingly adopted in areas such as decision-making \cite{liu2024grounded}, code generation \cite{yuan2024transagent}, and automated evaluation \cite{lifshitz2025multi}. More related work on MLLMs can be found in Appendix \ref{app:MLLM}.

\section{Methodology}

Figure \ref{fig:roadmap} illustrates the overall architecture of our multi-agent AES framework, which consists of \textbf{three core agents:} \ding{182} Initial Scorer, \ding{183} Feedback Pool Manager, and \ding{184} Reflective Scorer. To execute these agents, we introduce \textbf{two types of models}: \ding{182} a student MLLM and \ding{183} a teacher MLLM.
The student model, which has relatively weaker capabilities, handles the Initial Scorer by giving an initial score for each of the ten fine-grained traits. The teacher model, with stronger \textbf{reasoning abilities}, executes the Feedback Pool Manager to generate positive feedback comments and applies the Reflective Scorer to revise the student model’s initial scores based on the feedback  pool. This collaborative setup mirrors the human-in-the-loop process of "scoring → feedback → revision" in real-world scoring assessment. In the following sections, we describe each module in detail.

\subsection{Initial Scorer}
The Initial Scorer module is responsible for producing preliminary scores across the ten fine-grained traits. Given the text of the essay topic $T$, the corresponding image $I$, the student's essay $E$, and the detailed scoring rubrics $\bm{R}\in\mathbb{R}^{10}$, the student MLLM assigns an initial score \(s_i^{(0)}\) for each trait \(d_i\).
Formally, the Initial Scorer defines a mapping:
\begin{equation*}
\mathcal{S}: (T, I, E; \bm{R}) \mapsto \bm{s^{(0)}}\in\mathbb{R}^{10} 
\end{equation*}
where \(\bm{s^{(0)}} = (s_1^{(0)}, \ldots, s_{10}^{(0)})\) denotes the preliminary scores with \(s_i^{(0)} \in \{0,1,\ldots,5\}\).

This step can be viewed as the student model independently answering an exam based on its own understanding. The subsequent modules, executed by the teacher MLLM, are responsible for reviewing the student MLLMs' answers, providing feedback, and refining the initial judgments.

\subsection{Feedback Pool Manager}
The Feedback Pool Manager module is responsible for generating positive feedback for the student's essay based on the ten traits.  
Prior studies have indicated that MLLMs tend to adhere to the rubrics more strictly than human raters, often assigning lower scores during essay scoring \cite{su2025essayjudge, kundu2024largelanguagemodelsgood}. To address this tendency, we design the Feedback Pool Manager to focus exclusively on extracting positive feedback, emphasizing the strengths demonstrated in the essay. Formally, the Feedback Pool Manager defines a mapping:
\begin{equation*}
\mathcal{F}: (T, I, E; \bm{R}) \mapsto \bm{f}\in\mathbb{R}^{10}
\end{equation*}
where \(\bm{f} = (f_1, \ldots, f_{10})\) denotes a set of positive feedback entries, each associated with a specific trait \(d_i\).
For each trait, MLLMs return the extracted positive comments highlighting well-performed aspects of the essay.

The positive feedback generated by the teacher MLLM provides crucial and structured guidance for the Reflective Scorer in determining whether the initial scores require revision.

\subsection{Reflective Scorer}

The Reflective Scorer module is responsible for revising the student's initial scores by integrating positive feedback information.  
Formally, the Reflective Scorer defines a mapping:
\begin{equation*}
\mathcal{S^{'}}: (T, I, E, \bm{s^{(0)}}, \bm{f} ; \bm{R}) \mapsto \bm{s^{(1)}}\in\mathbb{R}^{10}
\end{equation*}
where \(\bm{s^{(1)}} = (s_1^{(1)}, \ldots, s_{10}^{(1)})\) denotes the revised scores across the ten traits. The teacher MLLM outputs a revised JSON object, and an example is shown in Figure \ref{fig:reflective_json_format}:

\begin{figure}[htbp]
  \centering
  \begin{tcolorbox}[
      colback=white,               
      colframe=black,              
      width=\linewidth,            
      enhanced jigsaw,             
      left=1mm, right=1mm,         
      top=1mm, bottom=1mm
    ]
    \textbf{\textit{revise}}: \textcolor{blue}{\textless\textbar} \textcolor{gray}{True / False}  \textcolor{blue}{\textbar\textgreater}

    \textbf{\textit{modified\_scores}}:

    \textcolor{blue}{\textless\textbar} Coherence:
    \textcolor{gray}{\{"score": X, "reason": "…"\}}\\
     Essay Length:
    \textcolor{gray}{\{"score": Y, "reason": "…"\}}
    \textcolor{blue}{\textbar\textgreater}
  \end{tcolorbox}
  \vspace{-4mm}
  \caption{Reflective scorer's JSON output format.}
  \label{fig:reflective_json_format}
\end{figure}

This reflective revision mechanism ensures that the final assessment fairly incorporates the strengths recognized in the essay, while avoiding unnecessary or overly aggressive adjustments.

\section{Experiments and Analysis}

\subsection{Experimental Setup}

\begin{table}[htbp]
\centering
\footnotesize
\renewcommand\tabcolsep{4pt}
\renewcommand\arraystretch{1.1}
\begin{tabular}{lc}
    \toprule
    \textbf{Statistic} & \textbf{Number} \\
    \midrule
    Total Multimodal Essays & 1,054 \\
    \midrule
    Image Type &  \\
    ~- Single-Image & 703 (66.7\%) \\
    ~- Multi-Image & 351 (33.3\%) \\
    \midrule
    Multimodal Essay Type &  \\
    ~- Flow Chart & 305 (28.9\%) \\
    ~- Bar Chart & 211 (20.0\%) \\
    ~- Table & 153 (14.5\%) \\
    ~- Line Chart & 145 (13.8\%) \\
    ~- Pie Chart & 71 (6.7\%) \\
    ~- Map & 62 (5.9\%) \\
    ~- Composite Chart & 107 (10.2\%) \\
    \bottomrule
\end{tabular}
\caption{Key statistics of the \textsc{EssayJudge} dataset.}
\label{tab:dataset_statistics}
\vspace{-2mm}
\end{table}

\textbf{Dataset.} We evaluate our agent-based AES system \agent on the \textsc{EssayJudge} dataset. It consists of 1,054 multimodal essays written at the university level. Each essay requires students to analyze and construct arguments based on visual inputs such as line charts and flow charts, posing significant challenges for MLLMs in terms of visual-textual understanding and reasoning. What's more, it covers 125 distinct essay topics across diverse domains including population, education, environment, production, and evolution. More details about the dataset are shown in Table \ref{tab:dataset_statistics}. The diversity in both topics and visual formats increases the complexity of the scoring task and provides a strong foundation for evaluating the robustness and generalizability of AES systems under varied multimodal scenarios.

\textbf{Basic Settings.} In the \agent AES framework, \textbf{GPT-4o is assigned as the default teacher model} throughout all experiments to guide and refine student MLLM's outputs, given its strong performance in AES \cite{openai2024gpt4ocard, su2025essayjudge}.
To evaluate \agent's generalization ability, we systematically assign a wide range of leading MLLMs to the student model, grouped as follows:
(i) \textbf{Open-source MLLMs}: InternVL2.5 (2B, 4B, 8B, 26B) \cite{internvl2.5}, Qwen2.5-VL (3B, 32B) \cite{qwen2_52025expanding}, and LLaMA-3.2-Vision (11B, 90B) \cite{grattafiori2024llama3herdmodels};
(ii) \textbf{Closed-source MLLMs}: Claude-3.5-Sonnet \cite{claude35s}, Gemini-2.5-Flash \cite{gemini}, and GPT-4o-mini \cite{openai2024gpt4omini}.
Since no existing AES model is designed for multimodal settings, we use the initial scores produced by each student model — before any teacher MLLM's feedback or reflection — to serve as the baseline for comparison. This setup ensures that any observed improvements can be fully attributed to the multi-agent process introduced in the \agent framework.
Detailed rubrics, prompts and model sources are listed in Appendix~\ref{app:rubrics}, \ref{app:prompt}, and~\ref{app:sources}.

\textbf{Evaluation Metric.} After extensively reviewing previous AES studies \cite{song2024automated, leeetal2024unleashing, mathias2018asap++, cozmaetal2018automated, wangetal2018automatic}, we select QWK as our evaluation metric, which is the most commonly used for assessing model alignment with ground truth scores. Its formula is expressed as:
\vspace{-1mm}
\[
k = 1 - \frac{\sum_{i,j}w_{i,j}O_{i,j}}{\sum_{i,j}w_{i,j}E_{i,j}},
\]
where $w_{i,j}=\frac{(i - j)^2}{\left(N - 1\right)^2}$ is the element of the weight matrix penalizing larger differences between $i$ and $j$, $\boldsymbol{O}_{i,j}$ is the observed agreement, and $\boldsymbol{E}_{i,j}$ is the expected agreement under random chance. QWK values range from -1 (complete disagreement) to 1 (perfect agreement). Higher values are expected. 

\subsection{Main Results}
\label{main results}
\begin{table*}[htb!]
\vspace{-3mm}
\centering
\small
\renewcommand\tabcolsep{3pt} 
\renewcommand\arraystretch{1.1} 
\resizebox{1.0\linewidth}{!}{
    \begin{tabular}{
        @{}
        c 
        | >{\centering\arraybackslash}m{0.8cm} 
        >{\centering\arraybackslash}m{0.8cm} 
        | >{\centering\arraybackslash}m{0.8cm} 
        >{\centering\arraybackslash}m{0.8cm} 
        >{\centering\arraybackslash}m{0.8cm} 
        >{\centering\arraybackslash}m{0.8cm} 
        | >{\centering\arraybackslash}m{0.8cm} 
        >{\centering\arraybackslash}m{0.8cm} 
        >{\centering\arraybackslash}m{0.8cm} 
        >{\centering\arraybackslash}m{0.8cm} 
        @{}
    }
    \toprule
    \multirow{2}{*}{\textbf{MLLMs}} &
    \multicolumn{2}{c|}{\textbf{Lexical Level}} &
    \multicolumn{4}{c|}{\textbf{Sentence Level}} &
    \multicolumn{4}{c}{\textbf{Discourse Level}} \\ 
    \cmidrule(lr){2-3} \cmidrule(lr){4-7} \cmidrule(lr){8-11}
     & \textbf{LA} & \textbf{LD} 
     & \textbf{CH} & \textbf{GA} & \textbf{GD} & \textbf{PA} 
     & \textbf{AC} & \textbf{JP} & \textbf{OS} & \textbf{EL} \\
    \midrule
    \multicolumn{11}{c}{\textit{\textbf{Open-Source MLLMs}}} \\
    \midrule
    
    {InternVL2.5-2B} \cite{internvl2.5} & 0.04 & 0.06 & 0.07 & 0.02 & 0.08 & 0.05 & 0.03 & 0.01 & 0.05 & 0.04\\
    \hfill~\textbf{+ \agent (Ours)} & \textbf{0.11} & \textbf{0.16} & \textbf{0.18}& \textbf{0.15} & \textbf{0.20} & \textbf{0.16} & \textbf{0.12} &\textbf{0.08} & \textbf{0.13} &\textbf{0.17} \\
    \hfill~\textbf{Improvements}&  \textbf{\up{0.07}} & \textbf{\up{0.10}} & \textbf{\up{0.11}} & \textbf{\up{0.13}} & \textbf{\up{0.12}} & \textbf{\up{0.11}} & \textbf{\up{0.10}} & \textbf{\up{0.07}} & \textbf{\up{0.08}} & \textbf{\up{0.13}} \\
    \midrule

    {InternVL2.5-4B} \cite{internvl2.5} & 0.12 & 0.14 & 0.35 & 0.08 & 0.11 & 0.13 & 0.24 & \textbf{0.29} & 0.37 & 0.30\\
    \hfill~\textbf{+ \agent (Ours)} & \textbf{0.23} & \textbf{0.37} & \textbf{0.40}& \textbf{0.31} & \textbf{0.35} & \textbf{0.27} & \textbf{0.24} & 0.24 & \textbf{0.33} &\textbf{0.34} \\
    \hfill~\textbf{Improvements}&  \textbf{\up{0.11}} & \textbf{\up{0.24}} & \textbf{\up{0.04}} & \textbf{\up{0.23}} & \textbf{\up{0.24}} & \textbf{\up{0.14}} & - & \down{0.04} & \textbf{\up{0.04}} & \textbf{\up{0.04}} \\
    \midrule
    
    {InternVL2.5-8B} \cite{internvl2.5} & 0.32 & 0.20 & 0.27 & 0.27 & 0.11 & 0.21 & 0.26 & 0.39 & 0.09 & 0.10\\
    \hfill~\textbf{+ \agent (Ours)} & \textbf{0.38} &\textbf{0.37} & \textbf{0.32} &\textbf{0.38} & \textbf{0.29} & \textbf{0.29} & \textbf{0.27} & \textbf{0.39}& \textbf{0.29} & \textbf{0.17} \\
    \hfill~\textbf{Improvements}& \textbf{\up{0.07}} & \textbf{\up{0.18}} & \textbf{\up{0.05}} & \textbf{\up{0.12}} & \textbf{\up{0.18}} & \textbf{\up{0.09}} & \textbf{\up{0.01}} & - & \textbf{\up{0.20}} & \textbf{\up{0.07}} \\
    \midrule
    
    {InternVL2.5-26B} \cite{internvl2.5} & \textbf{0.48} & 0.26 & 0.28 & \textbf{0.46} & 0.23 & \textbf{0.33} & \textbf{0.31} & \textbf{0.33} & 0.32 & 0.30\\
    \hfill~\textbf~\textbf{+ \agent (Ours)} & 0.42 & \textbf{0.38} & \textbf{0.30} & 0.40 & \textbf{0.35} & 0.31 & 0.25 & 0.29 & \textbf{0.34} & \textbf{0.31}  \\
    \hfill~\textbf{Improvements}& \down{0.06} & \textbf{\up{0.12}} & \textbf{\up{0.02}} & \down{0.05} & \textbf{\up{0.12}}& \down{0.02} &\down{0.07} & \down{0.04} & \textbf{\up{0.02}} & \textbf{\up{0.01}}\\
    
    \midrule
    
    {Qwen2.5-VL-3B} \cite{qwen2_52025expanding} & 0.19 & 0.28 & 0.34 & 0.19 & 0.29 & 0.20 & 0.26 & 0.29 & 0.34 & 0.32\\
    \hfill~\textbf{+ \agent (Ours)} &\textbf{0.30} &\textbf{0.30} &\textbf{0.39} & \textbf{0.44} &\textbf{0.32} &\textbf{0.34} & \textbf{0.27} &\textbf{ 0.35} & \textbf{0.37} & \textbf{0.35} \\
    \hfill~\textbf{Improvements}&  \textbf{\up{0.11}} & \textbf{\up{0.02}} & \textbf{\up{0.05}} & \textbf{\up{0.25}} & \textbf{\up{0.02}} & \textbf{\up{0.13}} & \textbf{\up{0.01}} & \textbf{\up{0.05}}& \textbf{\up{0.03}} & \textbf{\up{0.03}} \\
     
    \midrule
    
    {Qwen2.5-VL-32B} \cite{qwen2_52025expanding} & 0.43 & 0.40 & \textbf{0.50} & 0.48 & 0.39 & 0.38 &0.26 & 0.35 &\textbf{0.46} & 0.22\\
    \hfill~\textbf{+ \agent (Ours)} & \textbf{0.49} & \textbf{0.47} & 0.49 &\textbf{0.51} & \textbf{0.51} & \textbf{0.43} & \textbf{0.26}& \textbf{0.38} & 0.43 & \textbf{0.25}  \\
    \hfill~\textbf{Improvements}& \textbf{\up{0.06}} & \textbf{\up{0.07}} & \down{0.01} & \textbf{\up{0.03}} & \textbf{\up{0.13}} & \textbf{\up{0.05}}  & - & \textbf{\up{0.02}} & \down{0.03}  & \textbf{\up{0.03}} \\
    
    \midrule

    {LLaMA-3.2-Vision-11B} \cite{grattafiori2024llama3herdmodels} & 0.25 & 0.16 & 0.22 & 0.22 & 0.17 & 0.21 & 0.11 & 0.16 & 0.20 & 0.14\\
    \hfill~\textbf{+ \agent (Ours)} & \textbf{0.32} & \textbf{0.29} & \textbf{0.31} & \textbf{0.35} & \textbf{0.28} & \textbf{0.29} & \textbf{0.14} & \textbf{0.27}&\textbf{0.29} & \textbf{0.20}\\
    \hfill~\textbf{Improvements}& \textbf{\up{0.07}} & \textbf{\up{0.13}} & \textbf{\up{0.10}} & \textbf{\up{0.13}} & \textbf{\up{0.11}} & \textbf{\up{0.08}} & \textbf{\up{0.02}} & \textbf{\up{0.10}} & \textbf{\up{0.09}} & \textbf{\up{0.06}} \\

    \midrule
    
    {LLaMA-3.2-Vision-90B} \cite{grattafiori2024llama3herdmodels} & 0.40 & 0.29 & 0.38 & 0.40 & 0.30 & 0.32 & 0.21 & 0.30 & 0.35 & 0.16\\
    \hfill~\textbf{+ \agent (Ours)} & \textbf{0.45} &\textbf{0.42} & \textbf{0.43} & \textbf{0.46} &\textbf{0.44} & \textbf{0.39} & \textbf{0.25} &\textbf{ 0.33} & \textbf{0.37} & \textbf{0.22}  \\
    \hfill~\textbf{Improvements}& \textbf{\up{0.06}} & \textbf{\up{0.12}} & \textbf{\up{0.06}} & \textbf{\up{0.06}} & \textbf{\up{0.14}} & \textbf{\up{0.07}} & \textbf{\up{0.03}} & \textbf{\up{0.03}} & \textbf{\up{0.01}} & \textbf{\up{0.06}} \\

    \midrule
    \multicolumn{11}{c}{\textit{\textbf{Closed-Source MLLMs}}} \\
    \midrule

    {Claude-3.5-Sonnet} \cite{claude35s} & 0.50 & 0.53 & 0.49 & 0.54 & 0.52 & 0.55 & \textbf{0.41} & 0.39 & 0.57 & 0.27\\
    \hfill~\textbf{+ \agent (Ours)} & \textbf{0.58} & \textbf{0.59} & \textbf{0.55} & \textbf{0.55}& \textbf{0.58} &\textbf{0.55} & 0.39 & \textbf{0.45} & \textbf{0.57} & \textbf{0.35} \\
    \hfill~\textbf{Improvements}& \textbf{\up{0.08}} & \textbf{\up{0.06}} & \textbf{\up{0.07}} & \textbf{\up{0.01}} & \textbf{\up{0.06}} & \textbf{\up{0.01}}  & \down{0.02} & \textbf{\up{0.06}} & - & \textbf{\up{0.08}}  \\

    \midrule
    
    {Gemini-2.5-Flash} \cite{gemini} & 0.33 & 0.26 & 0.47 & 0.28 & 0.30 & 0.36 & \textbf{0.31} & 0.32 & \textbf{0.51} & 0.23\\
    \hfill~\textbf{+ \agent (Ours)} & \textbf{0.40}& \textbf{0.39} & \textbf{0.47} & \textbf{0.41} & \textbf{0.36} & \textbf{0.38} & 0.28 & \textbf{0.34} & 0.50 & \textbf{0.26} \\
    \hfill~\textbf{Improvements}& \textbf{\up{0.07}} & \textbf{\up{0.13}} & - & \textbf{\up{0.13}} & \textbf{\up{0.06}} & \textbf{\up{0.02}} & \down{0.03}  & \textbf{\up{0.03}} & \down{0.01}  & \textbf{\up{0.03}} \\ 	

    \midrule

    {GPT-4o-mini} \cite{openai2024gpt4omini} & 0.51 & 0.34 & 0.48 & \textbf{0.64} & 0.38 & \textbf{0.50} & 0.37 &\textbf{0.55} & 0.45 & 0.24\\
    \hfill~\textbf{+ \agent (Ours)} & \textbf{0.51} &\textbf{0.50} & \textbf{0.52} & 0.57 &\textbf{0.54} & 0.49 & \textbf{0.37} & 0.44 & \textbf{0.48} & \textbf{0.28}  \\
    \hfill~\textbf{Improvements}& - & \textbf{\up{0.16}} & \textbf{\up{0.04}}  & \down{0.07} & \textbf{\up{0.15}}  & \down{0.01}  & -  & \down{0.11}  & \textbf{\up{0.03}} & \textbf{\up{0.04}} \\

    \bottomrule
    \end{tabular}
}
\vspace{-2mm}
\caption{QWK scores of different student MLLMs on ten multi-granular essay traits. For each MLLM, the first row shows the baseline, the second shows the final result with \agent, and the third shows the improvement. Improvements are marked in \textbf{\textcolor{green2}{green}} arrow \textbf{\up{}} , while declines are indicated in \textbf{\textcolor{red2}{red}} arrow \textbf{\down{}}.}
\label{tab:main result}
\end{table*}

\textbf{Our proposed \agent framework yields consistent and significant improvements of QWK across each student MLLM on most traits.} Compared to the initial scores of single MLLMs, it improves the average QWK score by 0.07, from 0.29 to 0.36, representing a 21\% relative improvement. Notably, the Qwen2.5-VL-3B achieves a remarkable 0.25 QWK improvement on the Grammatical Accuracy trait, which is shown in Table \ref{tab:main result}). These results clearly demonstrate the robustness and effectiveness of our framework.

\begin{figure}[htb!]
\vspace{-2mm}
  \centering
  \includegraphics[width=0.48\textwidth]{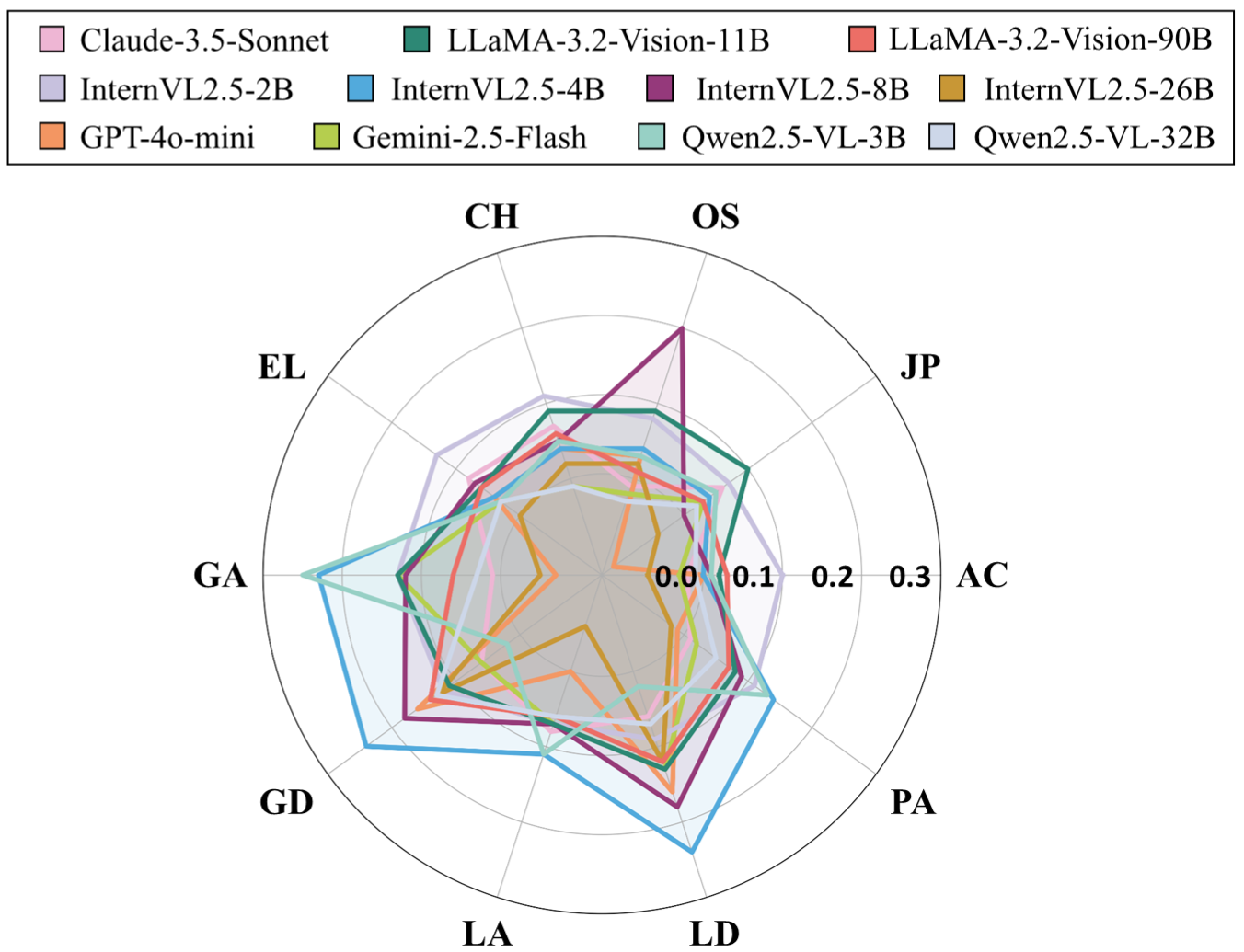}
  \caption{Trait-level score improvements after reflection via \agent across different student MLLMs.  }
  \label{fig:rador_improvements}
  \vspace{-2mm}
\end{figure}

\textbf{The \agent framework yields the most significant improvements in grammatical and lexical diversity.}
As shown in Figure~\ref{fig:rador_improvements}, these two traits show the largest improvements under the \agent framework compared to the initial scores. This is because single student MLLMs tend to focus on surface-level errors of grammar and vocabulary in initial scoring while overlooking the positive aspects of diversity \cite{su2025essayjudge, kundu2024largelanguagemodelsgood}, leading to underestimation compared to human raters (as shown in Appendix \ref{app:average score comparison}). With the help of the Feedback Pool Manager, the agent highlights strengths and passes them to the Reflective Scorer, enabling better recognition of diverse expression and more aligned score revisions.

\begin{figure*}[htb!]
    \centering
    \begin{minipage}{0.45\linewidth}
        \centering
        \includegraphics[width=\linewidth]{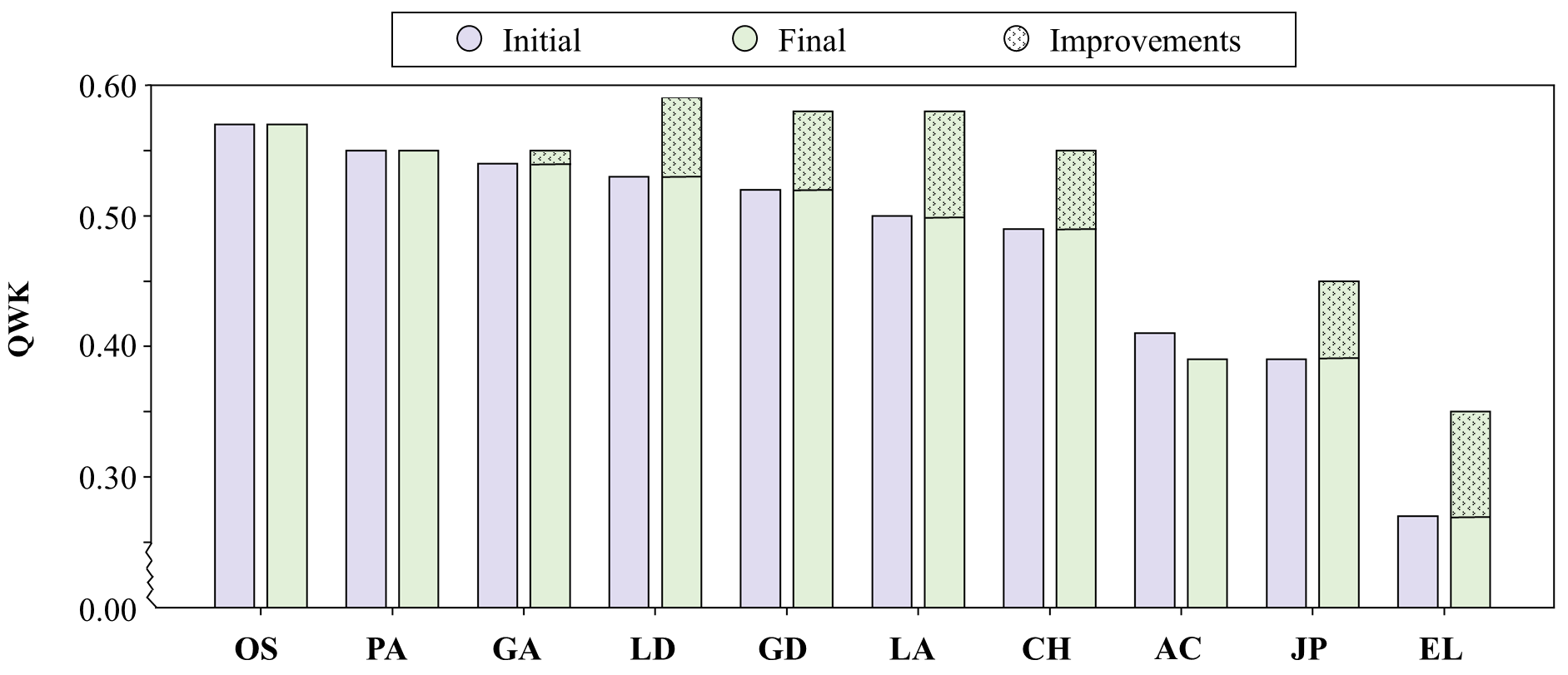}
        \caption*{(a) Claude-3.5-Sonnet}
    \end{minipage}
    \hspace{0.05\linewidth}
    \begin{minipage}{0.45\linewidth}
        \centering
        \includegraphics[width=\linewidth]{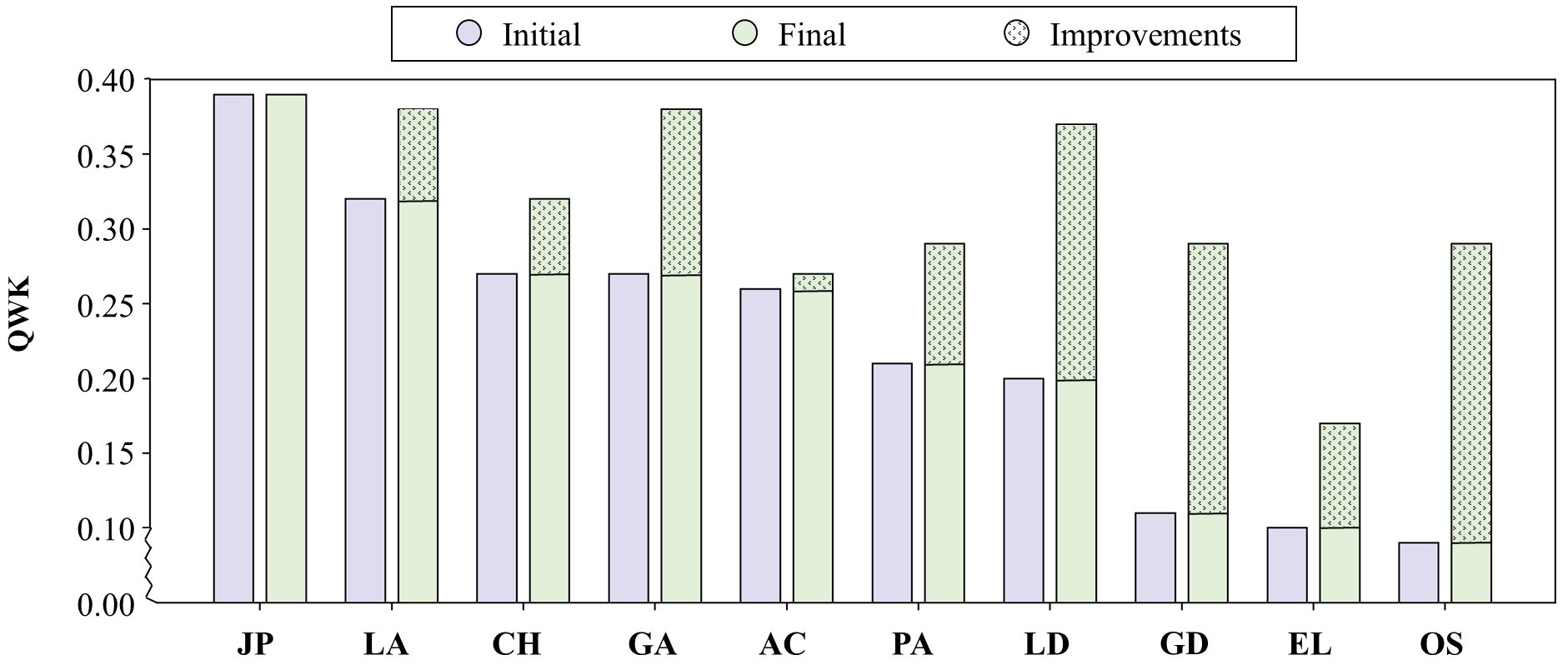}
        \caption*{(b) InternVL2.5-8B}
    \end{minipage}
    \caption{Improvements of QWK score across all traits based on different student MLLMs.}
    \label{fig:single_model_case}
\end{figure*}

\begin{figure}[t!]
  \centering
  \includegraphics[width=0.48\textwidth]{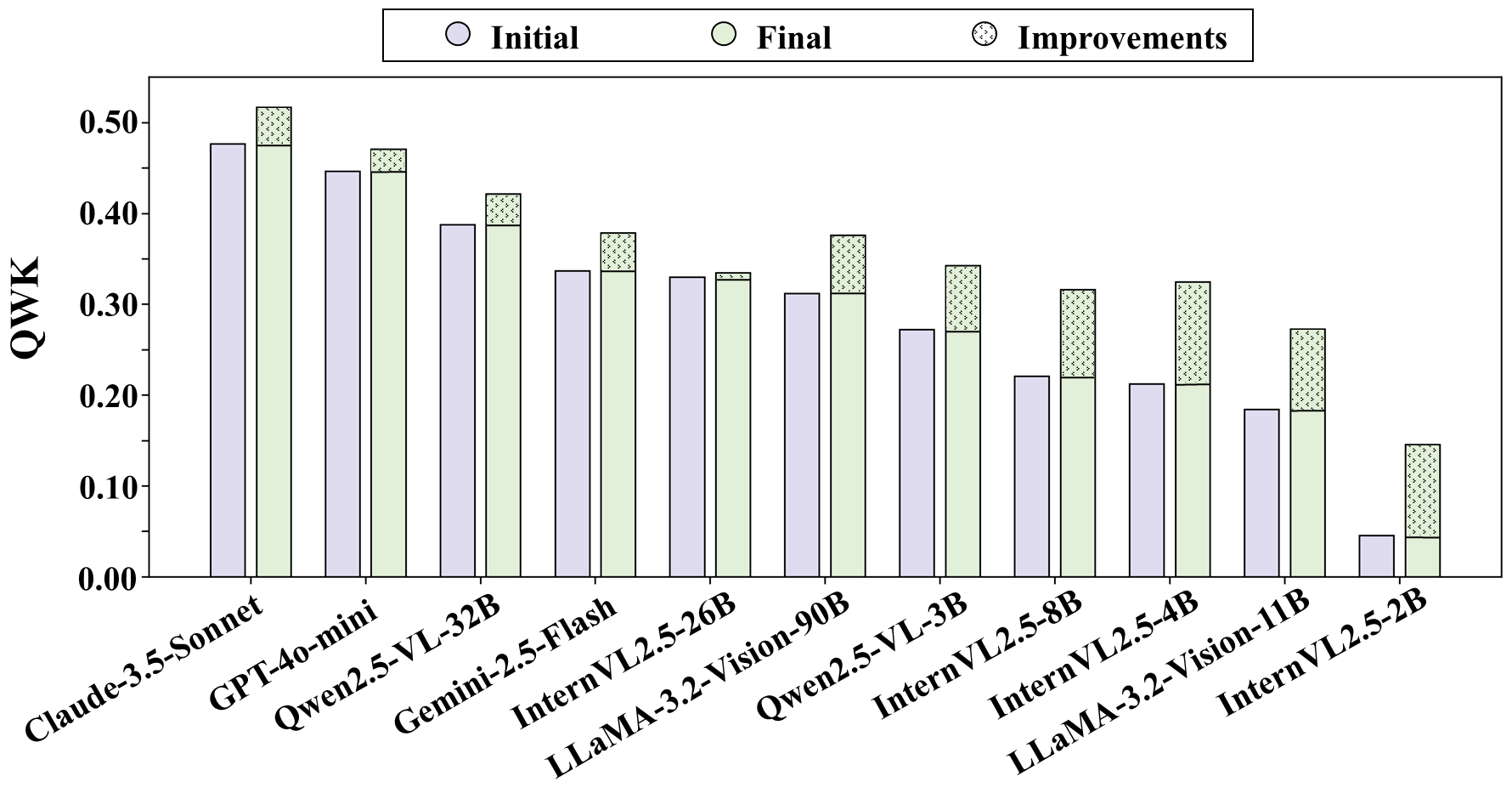}
  \caption{Improvements of average QWK score across ten traits of all student MLLMs.}
  \label{fig:all_models_comparison}
  \vspace{-2mm}
\end{figure}

\textbf{In general, the lower the QWK of initial score generated by Initial Scorer, the greater the QWK improvement brought by \agent.}
This trend appears both within and across student MLLMs. Within a student MLLM, traits with lower initial QWK tend to improve more with \agent (as shown in Figure~\ref{fig:single_model_case}). More examples are shown in Appendix \ref{app:more_examples_comparison}.  Across different student MLLMs, those with weaker initial performance benefit more from \agent framework, which is clearly demonstrated in Figure~\ref{fig:all_models_comparison}). This is likely because lower-performing traits or MLLMs have more room for improvement, while stronger ones are already close to the teacher MLLM’s level. 

\subsection{Analysis of \#image Setting}

\begin{figure}[htb!]
  \vspace{-2mm}
  \centering
  \includegraphics[width=0.48\textwidth]{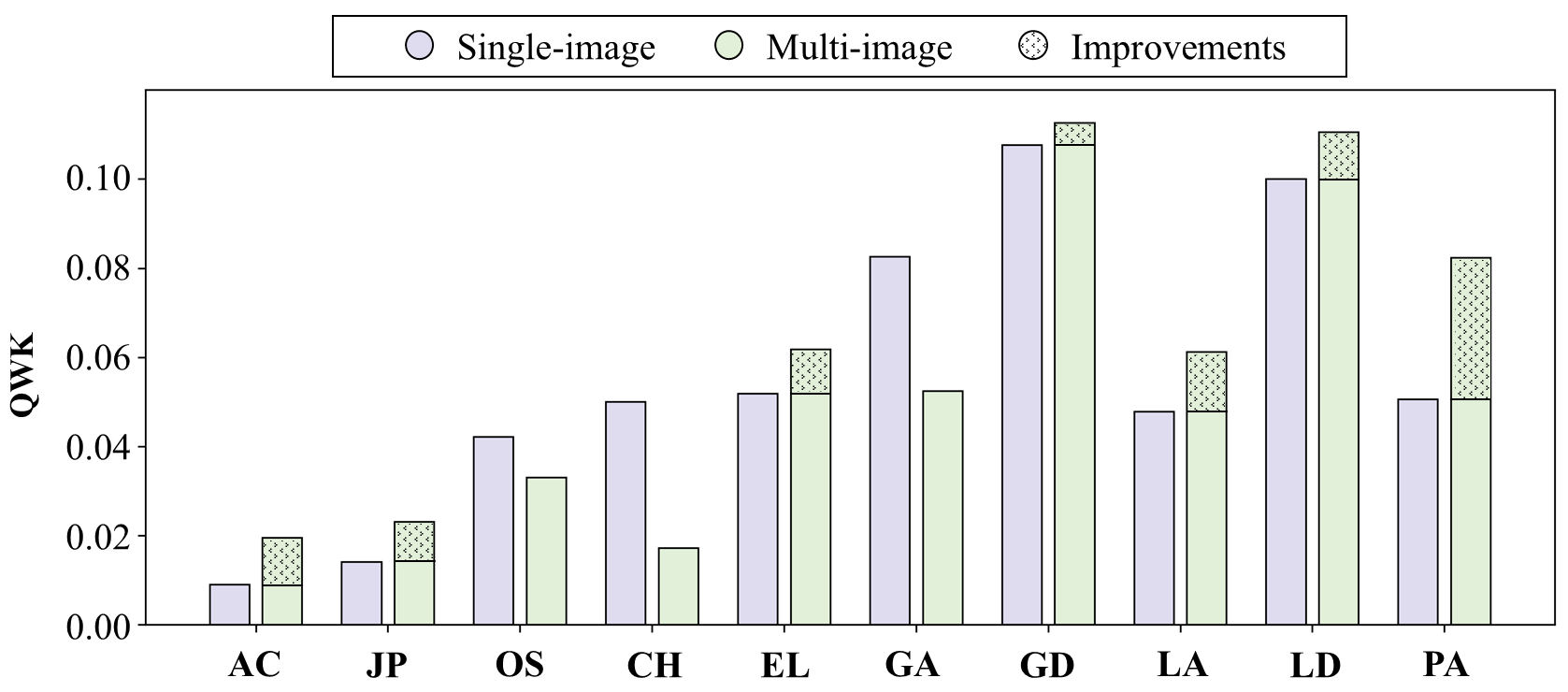}
  \caption{Improvements of average QWK of all student MLLMs under single-image and multi-image settings.}
  \label{fig:single_multi}
  \vspace{-4mm}
\end{figure}

\textbf{\agent framework yields greater improvements in the multi-image setting across most traits.} As shown in Figure ~\ref{fig:single_multi}, initial scores under multi-image settings tend to be more conservative, as MLLMs face greater difficulty in interpreting complex visual inputs. This conservative scoring provides more room for adjustment, allowing the \agent framework to achieve more noticeable improvements. Notably, these more improvements of QWK in multi-image settings also supports the necessity of incorporating multimodal inputs.

\textbf{For traits like Organizational Structure and Coherence, single-image topics yield greater improvements than multi-image topics.}
Unlike most traits where multi-image topics lead to the greatest improvements, organizational structure and coherence exhibit higher QWK improvements in the single-image setting (as shown in Figure \ref{fig:single_multi}). This may be attributed to the fact that single-image topics typically provide less visual information, which lowers the demand for essay structure and coherence in students' essays. In such cases, models find it easier to evaluate structural clarity and coherence, resulting in more confident scoring and greater improvements after reflection.

\begin{figure}[t!]
  \centering
  \includegraphics[width=0.48\textwidth]{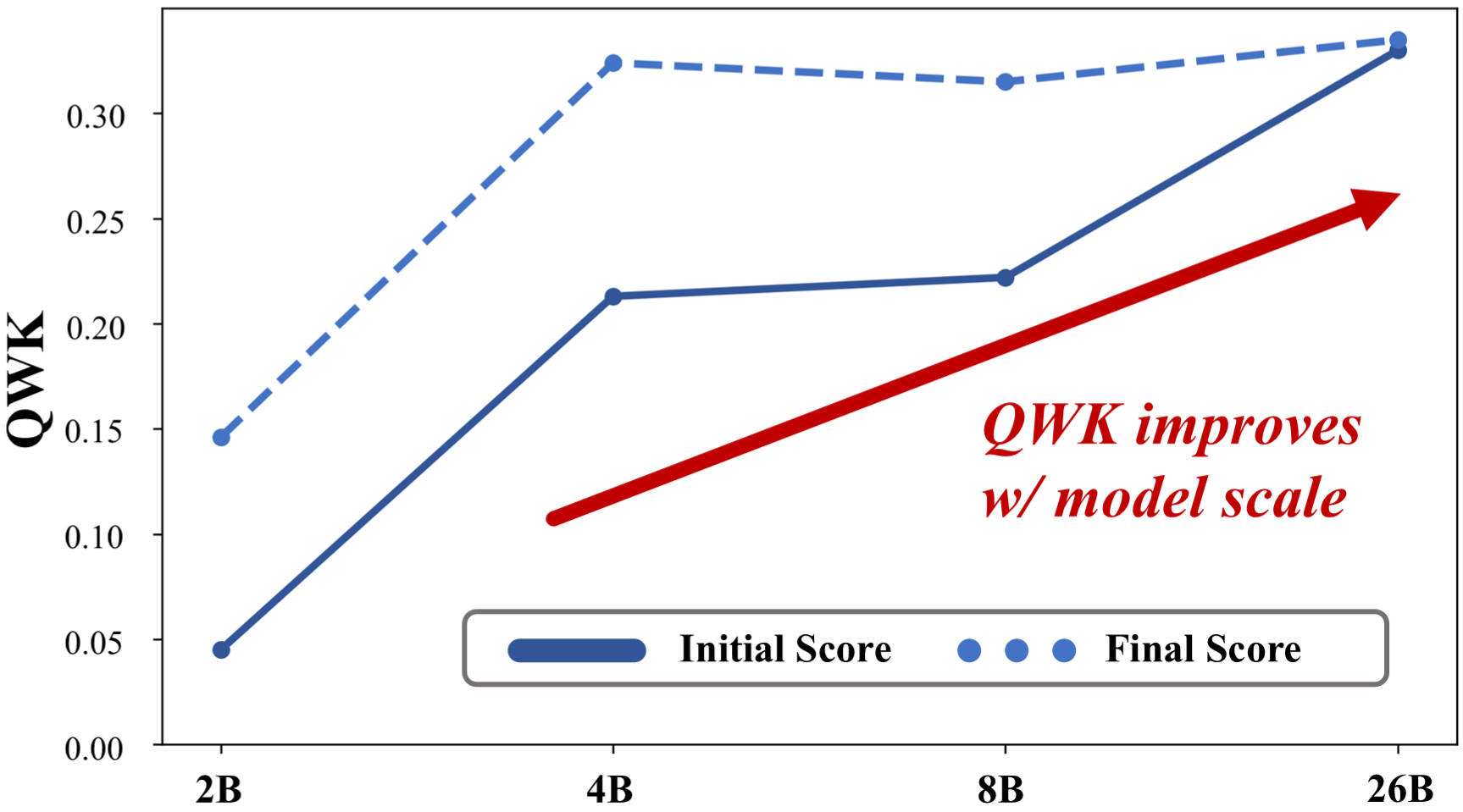}
  \caption{The average QWK scores of InternVL2.5 at different parameter scales (2B, 4B, 8B, 26B), evaluated before and after reflective feedback.}
  \label{fig:scaling}
  \vspace{-2mm}
\end{figure}

\subsection{Scaling Analysis}
\textbf{The performance of the student MLLM consistently improves with the scale of MLLM parameters.} We observe a trend similar to the scaling law \cite{kaplan2020scaling} in our setting. As shown in Figure \ref{fig:scaling}, when the size of InternVL2.5 increases from 2B to 26B, the average QWK score rises from 0.045 to 0.33 in the initial scoring stage. After incorporating \agent framework, the performance further improves, with the QWK increasing from 0.146 to 0.335. This result suggests that larger MLLMs exhibit stronger alignment with human judgment and greater reasoning ability.

\begin{figure}[ht!]
  \centering
  \includegraphics[width=0.48\textwidth]{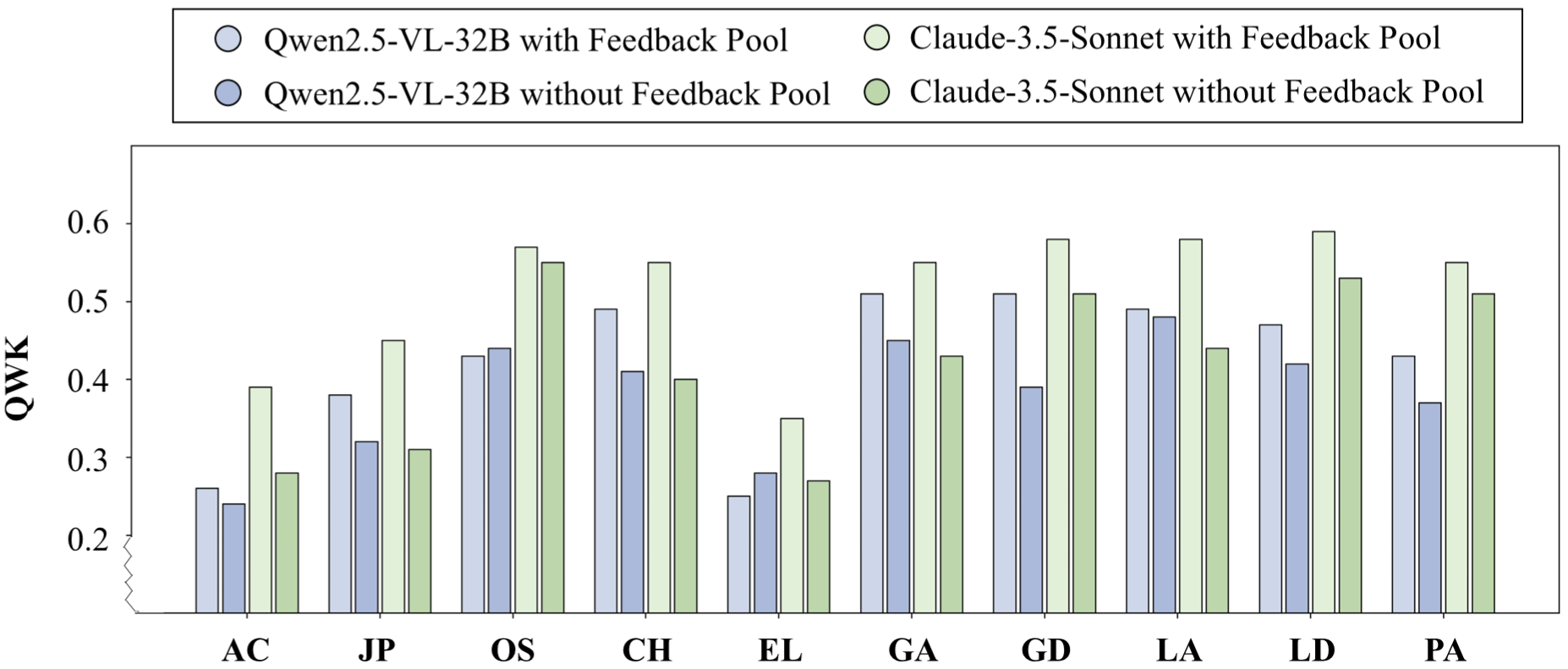}
  \caption{QWK changes w/ and w/o Feedback Pool for Qwen2.5-VL-32B \& Claude-3.5-Sonnet.}
  \label{fig:without_feedback}
  \vspace{-4mm}
\end{figure}

\begin{figure}[ht!]
  \centering
  \includegraphics[width=0.48\textwidth]{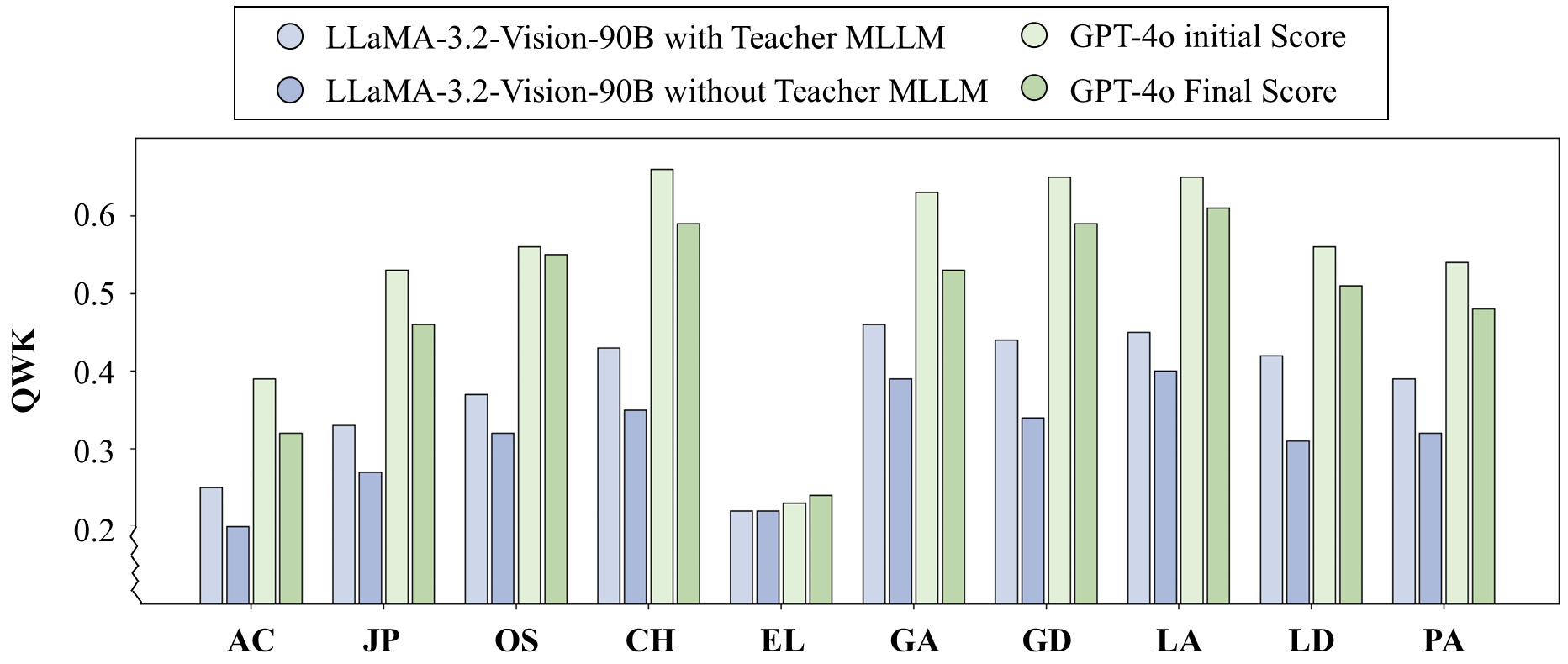}
  \caption{QWK scores across ten traits for two MLLMs (GPT-4o and LLaMA-3.2-Vision-90B), w/ and w/o the teacher-student collaboration mechanism. }
  \label{fig:without_teacher}
  \vspace{-4mm}
\end{figure}

\begin{figure*}[t!]
  \centering  \includegraphics[width=1\textwidth]{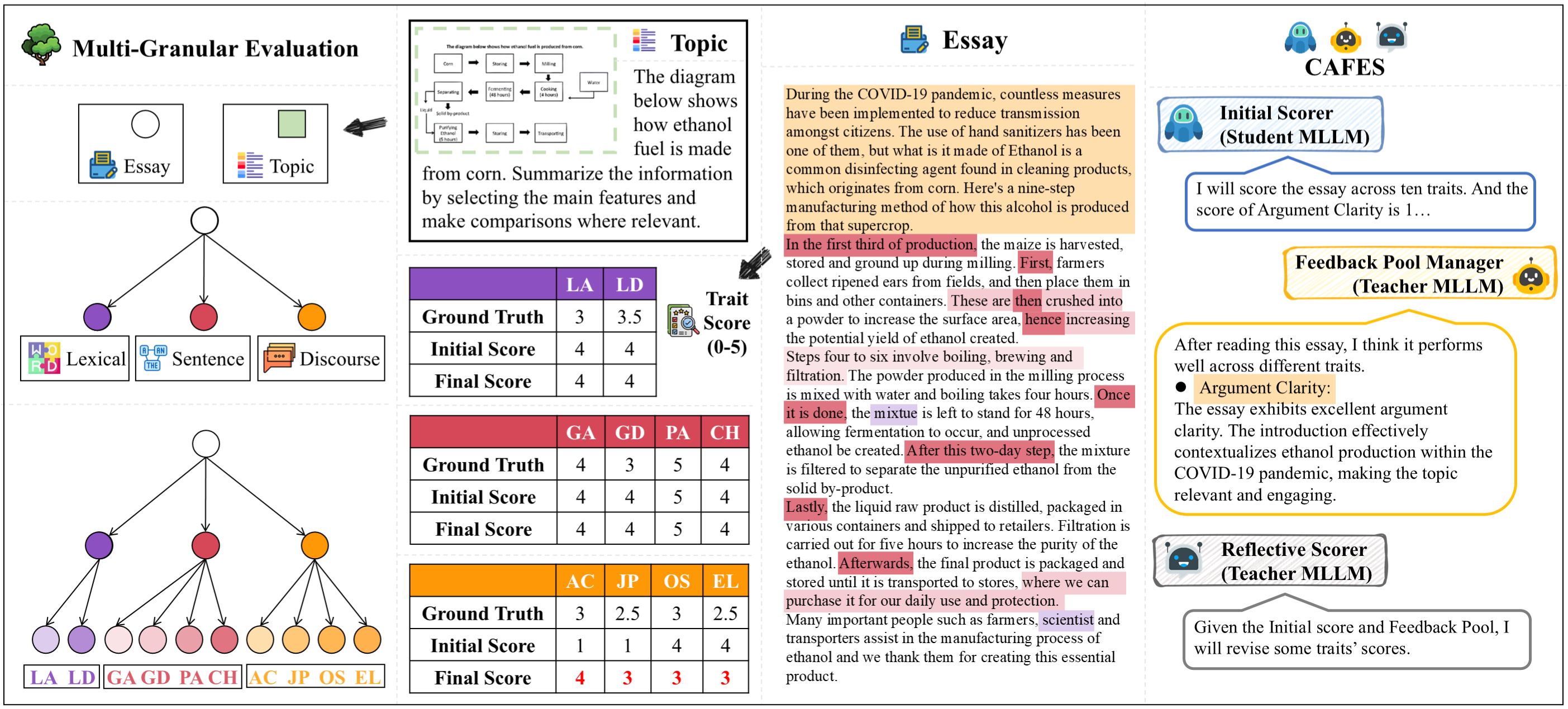}
  \caption{A representative case study illustrating \agent's score revision process. The student MLLM is Claude-3.5-Sonnet, and the teacher MLLM is GPT-4o.}
  \label{fig:case_study}
  \vspace{-4mm}
\end{figure*}

\subsection{Ablation Study}

We conduct two ablation studies to test key components of \agent. The first removes the Feedback Pool Manager. The second uses the same MLLM for both student and teacher models.

\textbf{Removing the Feedback Pool Manager results in reflected scores worse than the initial ones.} We allowed the initial scores to be directly revised without positive feedback. We apply this setup to two strong MLLMs as the student model: Claude-3.5-Sonnet (closed-source) and Qwen2.5-VL-32B (open-source). In both cases, QWK scores drop after reflection without feedback pool (Figure~\ref{fig:without_feedback}). This may be because \agent without positive feedback tends to over-focus on errors while ignoring strengths in essays, which underscores the importance of structured, trait-level positive feedback.

\textbf{Removing the teacher-student collaboration mechanism leads to a substantial drop in QWK.} We test two variants: using LLaMA-3.2-Vision-90B or GPT-4o for both student and teacher roles. As shown in Figure~\ref{fig:without_teacher}, QWK decreases notably compared to the original teacher–student setup, even when both roles use GPT-4o. These results suggest that different role assignment and independent reasoning between student and teacher MLLMs are essential for effective score revision in \agent’s cross-agent collaborative framework.

\subsection{Case Study}
To demonstrate how our \agent framework revises scores through feedback and reflection, we show an example using Claude-3.5-Sonnet as the student MLLM (as shown in Figure \ref{fig:case_study}). More examples are shown in Appendix \ref{app:case_study}. The essay explains how ethanol is produced, based on a flow chart. For example, we can find that the student MLLM initially gives a low argument clarity score of 1, while the ground truth score is 3. As mentioned in \ref{main results}, this is because the MLLM focuses too much on surface-level errors and overlooks key strengths, such as the relevant introduction and logical structure. After receiving positive, trait-specific feedback from the teacher MLLM, the Reflective Scorer revises the score upward. The final score better matches the human judgment, showing that targeted feedback helps correct overly harsh assessments and highlight overlooked merits.
This case shows how the \agent framework uses structured feedback and subsequent reflection to refine the initial model output, leading to better alignment with human scores and preferences.

\section{Conclusion}
\vspace{-2mm}
In this work, we present \agent, the first collaborative multi-agent framework for AES task. It divides the essay scoring process into three core stages (\textit{i.e.,} initial scoring, feedback generation, and reflective revision), enabling structured collaboration between three agents. Experiments across different student MLLMs show significant QWK improvements with \agent framework, especially in grammatical and lexical diversity. Ablation studies further confirm the necessity of the Feedback Pool Manager and teacher-student collaboration mechanism. We hope \agent can offer a new paradigm for building reliable and human-aligned AES systems and encourage the community to advance more effective and accurate scoring methods.

\clearpage

\section*{Limitations}
Despite the improvements we demonstrate in our \agent framework, there are still minor limitations:

\begin{enumerate}
    \item Our framework is evaluated on the EssayJudge dataset and achieves notable improvements over baseline models. However, EssayJudge — the only available multimodal essay dataset — focuses mainly on chart-based topics and does not cover more complex visual inputs such as film frames. We plan to include a broader range of multimodal essay types in future evaluations.
    \item The reflection mechanism helps suppress hallucinations from a single MLLM, such as fabricated justifications or misinterpretation of charts, but hallucination-induced scoring errors still occur. In future work, we aim to further strengthen evidence grounding.
\end{enumerate}




\bibliography{latex/cafes}

\begin{thebibliography}{88}
\providecommand{\natexlab}[1]{#1}

\bibitem[{Anthropic(2024)}]{claude35s}
Anthropic. 2024.
\newblock \href {https://www.anthropic.com/news/claude-3-5-sonnet} {Claude 3.5 sonnet}.

\bibitem[{Atkinson and Palma(2025)}]{atkinson2025llm}
John Atkinson and Diego Palma. 2025.
\newblock An llm-based hybrid approach for enhanced automated essay scoring.
\newblock \emph{Scientific Reports}, 15(1):14551.

\bibitem[{Attali and Burstein(2006)}]{Attali2006erator}
Yigal Attali and Jill Burstein. 2006.
\newblock Automated essay scoring with e-rater® v.2. journal of technology, learning, and assessment, 4(3).
\newblock \emph{Journal of Technology, Learning, and Assessment}, 4.

\bibitem[{Bailey and Meurers(2008)}]{bailey2008CREE}
Stacey Bailey and Detmar Meurers. 2008.
\newblock Diagnosing meaning errors in short answers to reading comprehension questions.
\newblock In \emph{Proceedings of the Third Workshop on Innovative Use of {NLP} for Building Educational Applications}, pages 107--115.

\bibitem[{Cai et~al.(2025)Cai, Liang, Lee, Wang, and Wu}]{cai2025rank}
Yida Cai, Kun Liang, Sanwoo Lee, Qinghan Wang, and Yunfang Wu. 2025.
\newblock Rank-then-score: Enhancing large language models for automated essay scoring.
\newblock \emph{arXiv preprint arXiv:2504.05736}.

\bibitem[{Cao et~al.(2020)Cao, Jin, Wan, and Yu}]{cao2020domain}
Yue Cao, Hanqi Jin, Xiaojun Wan, and Zhiwei Yu. 2020.
\newblock Domain-adaptive neural automated essay scoring.
\newblock In \emph{Proceedings of the 43rd International ACM SIGIR Conference on Research and Development in Information Retrieval}, SIGIR '20, page 1011–1020. Association for Computing Machinery.

\bibitem[{Chang et~al.(2024)Chang, Wang, Wang, Wu, Yang, Zhu, Chen, Yi, Wang, Wang et~al.}]{chang2024survey}
Yupeng Chang, Xu~Wang, Jindong Wang, Yuan Wu, Linyi Yang, Kaijie Zhu, Hao Chen, Xiaoyuan Yi, Cunxiang Wang, Yidong Wang, and 1 others. 2024.
\newblock A survey on evaluation of large language models.
\newblock \emph{ACM Transactions on Intelligent Systems and Technology}, 15(3):1--45.

\bibitem[{Chen and He(2013)}]{chen2013lexical}
Hongbo Chen and Ben He. 2013.
\newblock Automated essay scoring by maximizing human-machine agreement.
\newblock In \emph{Proceedings of the 2013 Conference on Empirical Methods in Natural Language Processing}, pages 1741--1752.

\bibitem[{Chen et~al.(2025{\natexlab{a}})Chen, Deng, Zheng, Yan, Liu, Wu, Jiang, Liu, and Hu}]{chen2025safeeraser}
Junkai Chen, Zhijie Deng, Kening Zheng, Yibo Yan, Shuliang Liu, PeiJun Wu, Peijie Jiang, Jia Liu, and Xuming Hu. 2025{\natexlab{a}}.
\newblock Safeeraser: Enhancing safety in multimodal large language models through multimodal machine unlearning.
\newblock \emph{arXiv preprint arXiv:2502.12520}.

\bibitem[{Chen et~al.(2025{\natexlab{b}})Chen, Wang, Cao, Liu, Gao, Cui, Zhu, Ye, Tian, Liu, Gu, Wang, Li, Ren, Chen, Luo, Wang, Jiang, Wang, He, Shi, Zhang, Lv, Wang, Shao, Chu, Tu, He, Wu, Deng, Ge, Chen, Zhang, Wang, Dou, Lu, Zhu, Lu, Lin, Qiao, Dai, and Wang}]{internvl2.5}
Zhe Chen, Weiyun Wang, Yue Cao, Yangzhou Liu, Zhangwei Gao, Erfei Cui, Jinguo Zhu, Shenglong Ye, Hao Tian, Zhaoyang Liu, Lixin Gu, Xuehui Wang, Qingyun Li, Yimin Ren, Zixuan Chen, Jiapeng Luo, Jiahao Wang, Tan Jiang, Bo~Wang, and 23 others. 2025{\natexlab{b}}.
\newblock \href {https://arxiv.org/abs/2412.05271} {Expanding performance boundaries of open-source multimodal models with model, data, and test-time scaling}.
\newblock \emph{Preprint}, arXiv:2412.05271.

\bibitem[{Chen et~al.(2025{\natexlab{c}})Chen, Wang, Cao, Liu, Gao, Cui, Zhu, Ye, Tian, Liu, Gu, Wang, Li, Ren, Chen, Luo, Wang, Jiang, Wang, He, Shi, Zhang, Lv, Wang, Shao, Chu, Tu, He, Wu, Deng, Ge, Chen, Zhang, Wang, Dou, Lu, Zhu, Lu, Lin, Qiao, Dai, and Wang}]{qwen2_52025expanding}
Zhe Chen, Weiyun Wang, Yue Cao, Yangzhou Liu, Zhangwei Gao, Erfei Cui, Jinguo Zhu, Shenglong Ye, Hao Tian, Zhaoyang Liu, Lixin Gu, Xuehui Wang, Qingyun Li, Yimin Ren, Zixuan Chen, Jiapeng Luo, Jiahao Wang, Tan Jiang, Bo~Wang, and 23 others. 2025{\natexlab{c}}.
\newblock \href {https://arxiv.org/abs/2412.05271} {Expanding performance boundaries of open-source multimodal models with model, data, and test-time scaling}.
\newblock \emph{Preprint}, arXiv:2412.05271.

\bibitem[{Choi et~al.(2025)Choi, Tate, Ritchie, Nixon, and Warschauer}]{choi2025anchor}
Jaeyoon Choi, Tamara Tate, Daniel Ritchie, Nia Nixon, and Mark Warschauer. 2025.
\newblock Anchor is the key: Toward accessible automated essay scoring with large language models through prompting.

\bibitem[{Chu et~al.(2025)Chu, Wang, Xie, Zhu, Yan, Ye, Zhong, Hu, Liang, Yu et~al.}]{chu2025llm}
Zhendong Chu, Shen Wang, Jian Xie, Tinghui Zhu, Yibo Yan, Jinheng Ye, Aoxiao Zhong, Xuming Hu, Jing Liang, Philip~S Yu, and 1 others. 2025.
\newblock Llm agents for education: Advances and applications.
\newblock \emph{arXiv preprint arXiv:2503.11733}.

\bibitem[{Cozma et~al.(2018{\natexlab{a}})Cozma, Butnaru, and Ionescu}]{cozma2018ASAP}
M{\u{a}}d{\u{a}}lina Cozma, Andrei Butnaru, and Radu~Tudor Ionescu. 2018{\natexlab{a}}.
\newblock Automated essay scoring with string kernels and word embeddings.
\newblock In \emph{Proceedings of the 56th Annual Meeting of the Association for Computational Linguistics (Volume 2: Short Papers)}, pages 503--509.

\bibitem[{Cozma et~al.(2018{\natexlab{b}})Cozma, Butnaru, and Ionescu}]{cozmaetal2018automated}
M{\u{a}}d{\u{a}}lina Cozma, Andrei Butnaru, and Radu~Tudor Ionescu. 2018{\natexlab{b}}.
\newblock \href {https://doi.org/10.18653/v1/P18-2080} {Automated essay scoring with string kernels and word embeddings}.
\newblock In \emph{Proceedings of the 56th Annual Meeting of the Association for Computational Linguistics (Volume 2: Short Papers)}, pages 503--509, Melbourne, Australia. Association for Computational Linguistics.

\bibitem[{Dang et~al.(2024)Dang, Huang, Huo, Yan, Huang, Liu, Gao, Zhang, Qian, Wang et~al.}]{dang2024explainable}
Yunkai Dang, Kaichen Huang, Jiahao Huo, Yibo Yan, Sirui Huang, Dongrui Liu, Mengxi Gao, Jie Zhang, Chen Qian, Kun Wang, and 1 others. 2024.
\newblock Explainable and interpretable multimodal large language models: A comprehensive survey.
\newblock \emph{arXiv preprint arXiv:2412.02104}.

\bibitem[{DeepMind(2025)}]{gemini}
Google DeepMind. 2025.
\newblock \href {https://deepmind.google/technologies/gemini/flash/} {Gemini 2.5 flash}.

\bibitem[{Dubey et~al.(2024)Dubey, Jauhri, Pandey, Kadian, Al-Dahle, Letman, Mathur, Schelten, Yang, Fan et~al.}]{grattafiori2024llama3herdmodels}
Abhimanyu Dubey, Abhinav Jauhri, Abhinav Pandey, Abhishek Kadian, Ahmad Al-Dahle, Aiesha Letman, Akhil Mathur, Alan Schelten, Amy Yang, Angela Fan, and 1 others. 2024.
\newblock The llama 3 herd of models.
\newblock \emph{arXiv preprint arXiv:2407.21783}.

\bibitem[{Granger et~al.(2009)Granger, Dagneaux, Meunier, and Paquot}]{granger2009icle}
Sylviane Granger, Estelle Dagneaux, Fanny Meunier, and Magali Paquot. 2009.
\newblock \emph{International Corpus of Learner English. Version 2. Handbook and CD-ROM}.

\bibitem[{Hu et~al.(2024)Hu, Tu, Han, He, Cui, Long, Zheng, Fang, Huang, Zhao, Zhang, Thai, Zhang, Wang, Yao, Zhao, Zhou, Cai, Zhai, Ding, Jia, Zeng, Li, Liu, and Sun}]{hu2024minicpmunveilingpotentialsmall}
Shengding Hu, Yuge Tu, Xu~Han, Chaoqun He, Ganqu Cui, Xiang Long, Zhi Zheng, Yewei Fang, Yuxiang Huang, Weilin Zhao, Xinrong Zhang, Zheng~Leng Thai, Kaihuo Zhang, Chongyi Wang, Yuan Yao, Chenyang Zhao, Jie Zhou, Jie Cai, Zhongwu Zhai, and 6 others. 2024.
\newblock \href {https://arxiv.org/abs/2404.06395} {Minicpm: Unveiling the potential of small language models with scalable training strategies}.
\newblock \emph{Preprint}, arXiv:2404.06395.

\bibitem[{Huang et~al.(2024)Huang, Huo, Yan, Wang, Yue, and Hu}]{huang2024miner}
Kaichen Huang, Jiahao Huo, Yibo Yan, Kun Wang, Yutao Yue, and Xuming Hu. 2024.
\newblock Miner: Mining the underlying pattern of modality-specific neurons in multimodal large language models.
\newblock \emph{arXiv preprint arXiv:2410.04819}.

\bibitem[{Huo et~al.(2024)Huo, Yan, Hu, Yue, and Hu}]{huo2024mmneuron}
Jiahao Huo, Yibo Yan, Boren Hu, Yutao Yue, and Xuming Hu. 2024.
\newblock Mmneuron: Discovering neuron-level domain-specific interpretation in multimodal large language model.
\newblock \emph{arXiv preprint arXiv:2406.11193}.

\bibitem[{Huo et~al.(2025)Huo, Yan, Zheng, Lyu, Zou, Wei, and Hu}]{huo2025mmunlearner}
Jiahao Huo, Yibo Yan, Xu~Zheng, Yuanhuiyi Lyu, Xin Zou, Zhihua Wei, and Xuming Hu. 2025.
\newblock Mmunlearner: Reformulating multimodal machine unlearning in the era of multimodal large language models.
\newblock \emph{arXiv preprint arXiv:2502.11051}.

\bibitem[{Hurst et~al.(2024)Hurst, Lerer, Goucher, Perelman, Ramesh, Clark, Ostrow, Welihinda, Hayes, Radford et~al.}]{openai2024gpt4ocard}
Aaron Hurst, Adam Lerer, Adam~P Goucher, Adam Perelman, Aditya Ramesh, Aidan Clark, AJ~Ostrow, Akila Welihinda, Alan Hayes, Alec Radford, and 1 others. 2024.
\newblock Gpt-4o system card.
\newblock \emph{arXiv preprint arXiv:2410.21276}.

\bibitem[{Jansen et~al.(2024)Jansen, Meyer, Fleckenstein, Horbach, Keller, and M{\"o}ller}]{jansen2024individualizing}
Thorben Jansen, Jennifer Meyer, Johanna Fleckenstein, Andrea Horbach, Stefan Keller, and Jens M{\"o}ller. 2024.
\newblock Individualizing goal-setting interventions using automated writing evaluation to support secondary school students’ text revisions.
\newblock \emph{Learning and Instruction}, 89:101847.

\bibitem[{Jiang et~al.(2023)Jiang, Gao, Yin, Liu, Yu, Cheng, and Gu}]{jiang2023improving}
Zhiwei Jiang, Tianyi Gao, Yafeng Yin, Meng Liu, Hua Yu, Zifeng Cheng, and Qing Gu. 2023.
\newblock \href {https://doi.org/10.18653/v1/2023.acl-long.696} {Improving domain generalization for prompt-aware essay scoring via disentangled representation learning}.
\newblock In \emph{Proceedings of the 61st Annual Meeting of the Association for Computational Linguistics (Volume 1: Long Papers)}, pages 12456--12470. Association for Computational Linguistics.

\bibitem[{Kamalov et~al.(2025)Kamalov, Calonge, Smail, Azizov, Thadani, Kwong, and Atif}]{kamalov2025evolution}
Firuz Kamalov, David~Santandreu Calonge, Linda Smail, Dilshod Azizov, Dimple~R Thadani, Theresa Kwong, and Amara Atif. 2025.
\newblock Evolution of ai in education: Agentic workflows.
\newblock \emph{arXiv preprint arXiv:2504.20082}.

\bibitem[{Kaplan et~al.(2020)Kaplan, McCandlish, Henighan, Brown, Chess, Child, Gray, Radford, Wu, and Amodei}]{kaplan2020scaling}
Jared Kaplan, Sam McCandlish, Tom Henighan, Tom~B Brown, Benjamin Chess, Rewon Child, Scott Gray, Alec Radford, Jeffrey Wu, and Dario Amodei. 2020.
\newblock Scaling laws for neural language models.
\newblock \emph{arXiv preprint arXiv:2001.08361}.

\bibitem[{Ke and Ng(2019)}]{Ke2019survey}
Zixuan Ke and Vincent Ng. 2019.
\newblock Automated essay scoring: A survey of the state of the art.
\newblock In \emph{Proceedings of the Twenty-Eighth International Joint Conference on Artificial Intelligence, {IJCAI-19}}, pages 6300--6308. International Joint Conferences on Artificial Intelligence Organization.

\bibitem[{Kundu and Barbosa(2024)}]{kundu2024largelanguagemodelsgood}
Anindita Kundu and Denilson Barbosa. 2024.
\newblock \href {https://arxiv.org/abs/2409.13120} {Are large language models good essay graders?}

\bibitem[{Lee et~al.(2024{\natexlab{a}})Lee, Cai, Meng, Wang, and Wu}]{lee2024TOEFL11}
Sanwoo Lee, Yida Cai, Desong Meng, Ziyang Wang, and Yunfang Wu. 2024{\natexlab{a}}.
\newblock Unleashing large language models' proficiency in zero-shot essay scoring.
\newblock In \emph{Findings of the Association for Computational Linguistics: EMNLP 2024}, pages 181--198.

\bibitem[{Lee et~al.(2024{\natexlab{b}})Lee, Cai, Meng, Wang, and Wu}]{leeetal2024unleashing}
Sanwoo Lee, Yida Cai, Desong Meng, Ziyang Wang, and Yunfang Wu. 2024{\natexlab{b}}.
\newblock \href {https://doi.org/10.18653/v1/2024.findings-emnlp.10} {Unleashing large language models' proficiency in zero-shot essay scoring}.
\newblock In \emph{Findings of the Association for Computational Linguistics: EMNLP 2024}, pages 181--198, Miami, Florida, USA. Association for Computational Linguistics.

\bibitem[{Li et~al.(2023)Li, Hammoud, Itani, Khizbullin, and Ghanem}]{li2023camelcommunicativeagentsmind}
Guohao Li, Hasan Abed Al~Kader Hammoud, Hani Itani, Dmitrii Khizbullin, and Bernard Ghanem. 2023.
\newblock \href {https://arxiv.org/abs/2303.17760} {Camel: Communicative agents for "mind" exploration of large language model society}.
\newblock \emph{Preprint}, arXiv:2303.17760.

\bibitem[{Li et~al.(2024)Li, Xu, Zhang, Chen, Liang, Fan, Li, Tang, and Wen}]{li2024bringing}
Hang Li, Tianlong Xu, Chaoli Zhang, Eason Chen, Jing Liang, Xing Fan, Haoyang Li, Jiliang Tang, and Qingsong Wen. 2024.
\newblock Bringing generative ai to adaptive learning in education.
\newblock \emph{arXiv preprint arXiv:2402.14601}.

\bibitem[{Li and Ng(2024{\natexlab{a}})}]{li2024reflection}
Shengjie Li and Vincent Ng. 2024{\natexlab{a}}.
\newblock Automated essay scoring: A reflection on the state of the art.
\newblock In \emph{Proceedings of the 2024 Conference on Empirical Methods in Natural Language Processing}, pages 17876--17888.

\bibitem[{Li and Ng(2024{\natexlab{b}})}]{li2024recent}
Shengjie Li and Vincent Ng. 2024{\natexlab{b}}.
\newblock Automated essay scoring: Recent successes and future directions.
\newblock In \emph{Proceedings of the Thirty-Third International Joint Conference on Artificial Intelligence, {IJCAI-24}}, pages 8114--8122.

\bibitem[{Li and Ng(2024{\natexlab{c}})}]{li2024icle++}
Shengjie Li and Vincent Ng. 2024{\natexlab{c}}.
\newblock Icle++: Modeling fine-grained traits for holistic essay scoring.
\newblock In \emph{Proceedings of the 2024 Conference of the North American Chapter of the Association for Computational Linguistics: Human Language Technologies (Volume 1: Long Papers)}, pages 8458--8478.

\bibitem[{Li and Liu(2024)}]{li2024applying}
Wenchao Li and Haitao Liu. 2024.
\newblock Applying large language models for automated essay scoring for non-native japanese.
\newblock \emph{Humanities and Social Sciences Communications}, 11(1):1--15.

\bibitem[{Liang et~al.(2024)Liang, He, Jiao, Wang, Wang, Wang, Yang, Shi, and Tu}]{liang2024encouraging}
Tian Liang, Zhiwei He, Wenxiang Jiao, Xing Wang, Yan Wang, Rui Wang, Yujiu Yang, Shuming Shi, and Zhaopeng Tu. 2024.
\newblock Encouraging divergent thinking in large language models through multi-agent debate.
\newblock In \emph{Proceedings of the 2024 Conference on Empirical Methods in Natural Language Processing}.

\bibitem[{Lifshitz et~al.(2025)Lifshitz, McIlraith, and Du}]{lifshitz2025multi}
Shalev Lifshitz, Sheila~A McIlraith, and Yilun Du. 2025.
\newblock Multi-agent verification: Scaling test-time compute with multiple verifiers.
\newblock \emph{arXiv preprint arXiv:2502.20379}.

\bibitem[{Lim et~al.(2021)Lim, Bong, Wong, and Lee}]{lim2021comprehensive}
Chun~Then Lim, Chih~How Bong, Wee~Sian Wong, and Nung~Kion Lee. 2021.
\newblock A comprehensive review of automated essay scoring (aes) research and development.
\newblock \emph{Pertanika Journal of Science \& Technology}, 29(3):1875--1899.

\bibitem[{Liu et~al.(2024{\natexlab{a}})Liu, Li, Li, Li, Zhang, Shen, and Lee}]{liu2024llavanext}
Haotian Liu, Chunyuan Li, Yuheng Li, Bo~Li, Yuanhan Zhang, Sheng Shen, and Yong~Jae Lee. 2024{\natexlab{a}}.
\newblock \href {https://llava-vl.github.io/blog/2024-01-30-llava-next/} {Llava-next: Improved reasoning, ocr, and world knowledge}.

\bibitem[{Liu et~al.(2024{\natexlab{b}})Liu, Yang, Sun, Qian, Wan, Chen, and Lan}]{liu2024grounded}
Zeyang Liu, Xinrui Yang, Shiguang Sun, Long Qian, Lipeng Wan, Xingyu Chen, and Xuguang Lan. 2024{\natexlab{b}}.
\newblock Grounded answers for multi-agent decision-making problem through generative world model.
\newblock \emph{Advances in Neural Information Processing Systems}, 37:46622--46652.

\bibitem[{Lu et~al.(2024)Lu, Liu, Zhang, Wang, Dong, Liu, Sun, Ren, Li, Sun et~al.}]{lu2024deepseek}
Haoyu Lu, Wen Liu, Bo~Zhang, Bingxuan Wang, Kai Dong, Bo~Liu, Jingxiang Sun, Tongzheng Ren, Zhuoshu Li, Yaofeng Sun, and 1 others. 2024.
\newblock Deepseek-vl: towards real-world vision-language understanding.
\newblock \emph{arXiv preprint arXiv:2403.05525}.

\bibitem[{Mansour et~al.(2024)Mansour, Albatarni, Eltanbouly, and Elsayed}]{mansour2024fewshoting}
Watheq Mansour, Salam Albatarni, Sohaila Eltanbouly, and Tamer Elsayed. 2024.
\newblock \href {https://arxiv.org/abs/2403.06149} {Can large language models automatically score proficiency of written essays?}
\newblock \emph{Preprint}, arXiv:2403.06149.

\bibitem[{Mathias and Bhattacharyya(2018)}]{mathias2018asap++}
Sandeep Mathias and Pushpak Bhattacharyya. 2018.
\newblock {ASAP}++: Enriching the {ASAP} automated essay grading dataset with essay attribute scores.
\newblock In \emph{Proceedings of the Eleventh International Conference on Language Resources and Evaluation ({LREC} 2018)}.

\bibitem[{Mizumoto and Eguchi(2023)}]{MIZUMOTO2023llmbased}
Atsushi Mizumoto and Masaki Eguchi. 2023.
\newblock Exploring the potential of using an ai language model for automated essay scoring.
\newblock \emph{Research Methods in Applied Linguistics}, 2(2):100050.

\bibitem[{OpenAI(2024)}]{openai2024gpt4omini}
OpenAI. 2024.
\newblock \href {https://openai.com/index/gpt-4o-mini-advancing-cost-efficient-intelligence/} {Gpt-4o mini: advancing cost-efficient intelligence}.

\bibitem[{Qu et~al.(2025)Qu, Dai, Wei, Cai, Wang, Yin, Xu, and Wen}]{qu2025tool}
Changle Qu, Sunhao Dai, Xiaochi Wei, Hengyi Cai, Shuaiqiang Wang, Dawei Yin, Jun Xu, and Ji-Rong Wen. 2025.
\newblock Tool learning with large language models: A survey.
\newblock \emph{Frontiers of Computer Science}, 19(8):198343.

\bibitem[{Ramesh and Sanampudi(2022)}]{ramesh2022automated}
Dadi Ramesh and Suresh~Kumar Sanampudi. 2022.
\newblock An automated essay scoring systems: a systematic literature review.
\newblock \emph{Artificial Intelligence Review}, 55(3):2495--2527.

\bibitem[{Song et~al.(2025{\natexlab{a}})Song, Yuk, Choi, Yoo, Lim, Lim, and Park}]{song2025unified}
SeungWoo Song, Junghun Yuk, ChangSu Choi, HanGyeol Yoo, Hyeonseok Lim, KyungTae Lim, and Jungyeul Park. 2025{\natexlab{a}}.
\newblock Unified automated essay scoring and grammatical error correction.
\newblock In \emph{Findings of the Association for Computational Linguistics: NAACL 2025}, pages 4412--4426.

\bibitem[{Song et~al.(2025{\natexlab{b}})Song, Li, Li, Zhao, Yu, Ma, Mao, Zhang, and Wang}]{song2025bridge}
Shezheng Song, Xiaopeng Li, Shasha Li, Shan Zhao, Jie Yu, Jun Ma, Xiaoguang Mao, Weimin Zhang, and Meng Wang. 2025{\natexlab{b}}.
\newblock How to bridge the gap between modalities: Survey on multimodal large language model.
\newblock \emph{IEEE Transactions on Knowledge and Data Engineering}.

\bibitem[{Song et~al.(2024)Song, Zhu, Wang, and Zheng}]{song2024automated}
Yishen Song, Qianta Zhu, Huaibo Wang, and Qinhua Zheng. 2024.
\newblock Automated essay scoring and revising based on open-source large language models.
\newblock \emph{IEEE Transactions on Learning Technologies}.

\bibitem[{Stab and Gurevych(2014)}]{stab2014AAE}
Christian Stab and Iryna Gurevych. 2014.
\newblock Annotating argument components and relations in persuasive essays.
\newblock In \emph{Proceedings of {COLING} 2014, the 25th International Conference on Computational Linguistics: Technical Papers}, pages 1501--1510.

\bibitem[{Su et~al.(2025)Su, Yan, Fu, Zhang, Ye, Liu, Huo, Zhou, and Hu}]{su2025essayjudge}
Jiamin Su, Yibo Yan, Fangteng Fu, Han Zhang, Jingheng Ye, Xiang Liu, Jiahao Huo, Huiyu Zhou, and Xuming Hu. 2025.
\newblock Essayjudge: A multi-granular benchmark for assessing automated essay scoring capabilities of multimodal large language models.

\bibitem[{Tran et~al.(2025)Tran, Dao, Nguyen, Pham, O'Sullivan, and Nguyen}]{tran2025multiagentcollaborationmechanismssurvey}
Khanh-Tung Tran, Dung Dao, Minh-Duong Nguyen, Quoc-Viet Pham, Barry O'Sullivan, and Hoang~D. Nguyen. 2025.
\newblock Multi-agent collaboration mechanisms: A survey of llms.

\bibitem[{Uto(2021)}]{uto2021review}
Masaki Uto. 2021.
\newblock A review of deep-neural automated essay scoring models.
\newblock \emph{Behaviormetrika}, 48(2):459--484.

\bibitem[{Uto et~al.(2020)Uto, Xie, and Ueno}]{uto2020neural}
Masaki Uto, Yikuan Xie, and Maomi Ueno. 2020.
\newblock Neural automated essay scoring incorporating handcrafted features.
\newblock In \emph{Proceedings of the 28th international conference on computational linguistics}, pages 6077--6088.

\bibitem[{Vajjala(2016)}]{Sowmya2016linguistic}
Sowmya Vajjala. 2016.
\newblock Automated assessment of non-native learner essays: Investigating the role of linguistic features.
\newblock \emph{CoRR}.

\bibitem[{Wang et~al.(2025)Wang, Zhang, Zhou, Wu, Yu, Zhao, Yin, Fu, Yan, Luo et~al.}]{wang2025comprehensive}
Kun Wang, Guibin Zhang, Zhenhong Zhou, Jiahao Wu, Miao Yu, Shiqian Zhao, Chenlong Yin, Jinhu Fu, Yibo Yan, Hanjun Luo, and 1 others. 2025.
\newblock A comprehensive survey in llm (-agent) full stack safety: Data, training and deployment.
\newblock \emph{arXiv preprint arXiv:2504.15585}.

\bibitem[{Wang et~al.(2022)Wang, Wang, Li, and Lin}]{wang2022bert}
Yongjie Wang, Chuang Wang, Ruobing Li, and Hui Lin. 2022.
\newblock On the use of bert for automated essay scoring: Joint learning of multi-scale essay representation.
\newblock In \emph{Proceedings of the 2022 Conference of the North American Chapter of the Association for Computational Linguistics: Human Language Technologies}, pages 3416--3425.

\bibitem[{Wang et~al.(2018)Wang, Wei, Zhou, and Huang}]{wangetal2018automatic}
Yucheng Wang, Zhongyu Wei, Yaqian Zhou, and Xuanjing Huang. 2018.
\newblock \href {https://doi.org/10.18653/v1/D18-1090} {Automatic essay scoring incorporating rating schema via reinforcement learning}.
\newblock In \emph{Proceedings of the 2018 Conference on Empirical Methods in Natural Language Processing}, pages 791--797, Brussels, Belgium. Association for Computational Linguistics.

\bibitem[{Wu et~al.(2023)Wu, Bansal, Zhang, Wu, Li, Zhu, Jiang, Zhang, Zhang, Liu, Awadallah, White, Burger, and Wang}]{wu2023autogenenablingnextgenllm}
Qingyun Wu, Gagan Bansal, Jieyu Zhang, Yiran Wu, Beibin Li, Erkang Zhu, Li~Jiang, Xiaoyun Zhang, Shaokun Zhang, Jiale Liu, Ahmed~Hassan Awadallah, Ryen~W White, Doug Burger, and Chi Wang. 2023.
\newblock \href {https://arxiv.org/abs/2308.08155} {Autogen: Enabling next-gen llm applications via multi-agent conversation}.
\newblock \emph{Preprint}, arXiv:2308.08155.

\bibitem[{Wu et~al.(2024)Wu, Saraf, Lee, Latif, Liu, and Zhai}]{wu2024unveiling}
Xuansheng Wu, Padmaja~Pravin Saraf, Gyeong-Geon Lee, Ehsan Latif, Ninghao Liu, and Xiaoming Zhai. 2024.
\newblock Unveiling scoring processes: Dissecting the differences between llms and human graders in automatic scoring.
\newblock \emph{arXiv preprint arXiv:2407.18328}.

\bibitem[{Xi et~al.(2023)Xi, Chen, Guo, He, Ding, Hong, Zhang, Wang, Jin, Zhou et~al.}]{xi2023rise}
Zhiheng Xi, Wenxiang Chen, Xin Guo, Wei He, Yiwen Ding, Boyang Hong, Ming Zhang, Junzhe Wang, Senjie Jin, Enyu Zhou, and 1 others. 2023.
\newblock The rise and potential of large language model based agents: A survey.
\newblock \emph{arXiv preprint arXiv:2309.07864}.

\bibitem[{Xia et~al.(2024)Xia, Mao, and Zheng}]{xia2024empirical}
Wei Xia, Shaoguang Mao, and Chanjing Zheng. 2024.
\newblock Empirical study of large language models as automated essay scoring tools in english composition\_taking toefl independent writing task for example.
\newblock \emph{arXiv preprint arXiv:2401.03401}.

\bibitem[{Xiao et~al.(2024)Xiao, Ma, Song, Xu, Zhang, Wang, and Fu}]{xiao2024fewshoting}
Changrong Xiao, Wenxing Ma, Qingping Song, Sean~Xin Xu, Kunpeng Zhang, Yufang Wang, and Qi~Fu. 2024.
\newblock \href {https://arxiv.org/abs/2401.06431} {Human-ai collaborative essay scoring: A dual-process framework with llms}.
\newblock \emph{Preprint}, arXiv:2401.06431.

\bibitem[{Xu et~al.(2025)Xu, Shahreeza, Hoo, and Yang}]{xu2025explainable}
Wenbo Xu, Muhammad Shahreeza, Wai~Lam Hoo, and Wudao Yang. 2025.
\newblock Explainable ai for education: Enhancing essay scoring via rubric-aligned chain-of-thought prompting.

\bibitem[{Yan and Lee(2024)}]{yan2024georeasoner}
Yibo Yan and Joey Lee. 2024.
\newblock Georeasoner: Reasoning on geospatially grounded context for natural language understanding.
\newblock In \emph{Proceedings of the 33rd ACM International Conference on Information and Knowledge Management}, pages 4163--4167.

\bibitem[{Yan et~al.(2024{\natexlab{a}})Yan, Su, He, Fu, Zheng, Lyu, Wang, Wang, Wen, and Hu}]{yan2024survey}
Yibo Yan, Jiamin Su, Jianxiang He, Fangteng Fu, Xu~Zheng, Yuanhuiyi Lyu, Kun Wang, Shen Wang, Qingsong Wen, and Xuming Hu. 2024{\natexlab{a}}.
\newblock A survey of mathematical reasoning in the era of multimodal large language model: Benchmark, method \& challenges.
\newblock \emph{arXiv preprint arXiv:2412.11936}.

\bibitem[{Yan et~al.(2024{\natexlab{b}})Yan, Wang, Huo, Li, Li, Su, Gao, Zhang, Xu, Chu et~al.}]{yan2024errorradar}
Yibo Yan, Shen Wang, Jiahao Huo, Hang Li, Boyan Li, Jiamin Su, Xiong Gao, Yi-Fan Zhang, Tianlong Xu, Zhendong Chu, and 1 others. 2024{\natexlab{b}}.
\newblock Errorradar: Benchmarking complex mathematical reasoning of multimodal large language models via error detection.
\newblock \emph{arXiv preprint arXiv:2410.04509}.

\bibitem[{Yan et~al.(2025{\natexlab{a}})Yan, Wang, Huo, Ye, Chu, Hu, Yu, Gomes, Selman, and Wen}]{yan2025position}
Yibo Yan, Shen Wang, Jiahao Huo, Jingheng Ye, Zhendong Chu, Xuming Hu, Philip~S Yu, Carla Gomes, Bart Selman, and Qingsong Wen. 2025{\natexlab{a}}.
\newblock Position: Multimodal large language models can significantly advance scientific reasoning.
\newblock \emph{arXiv preprint arXiv:2502.02871}.

\bibitem[{Yan et~al.(2025{\natexlab{b}})Yan, Wang, Huo, Yu, Hu, and Wen}]{yan2025mathagent}
Yibo Yan, Shen Wang, Jiahao Huo, Philip~S Yu, Xuming Hu, and Qingsong Wen. 2025{\natexlab{b}}.
\newblock Mathagent: Leveraging a mixture-of-math-agent framework for real-world multimodal mathematical error detection.
\newblock \emph{arXiv preprint arXiv:2503.18132}.

\bibitem[{Yan et~al.(2024{\natexlab{c}})Yan, Wen, Zhong, Chen, Chen, Wen, Zimmermann, and Liang}]{yan2024urbanclip}
Yibo Yan, Haomin Wen, Siru Zhong, Wei Chen, Haodong Chen, Qingsong Wen, Roger Zimmermann, and Yuxuan Liang. 2024{\natexlab{c}}.
\newblock Urbanclip: Learning text-enhanced urban region profiling with contrastive language-image pretraining from the web.
\newblock In \emph{Proceedings of the ACM on Web Conference 2024}, pages 4006--4017.

\bibitem[{Yang et~al.(2024)Yang, Rakovi{\'c}, Li, Guan, Ga{\v{s}}evi{\'c}, and Chen}]{yang2024unveiling}
Kaixun Yang, Mladen Rakovi{\'c}, Yuyang Li, Quanlong Guan, Dragan Ga{\v{s}}evi{\'c}, and Guangliang Chen. 2024.
\newblock Unveiling the tapestry of automated essay scoring: A comprehensive investigation of accuracy, fairness, and generalizability.
\newblock In \emph{Proceedings of the AAAI Conference on Artificial Intelligence}, volume~38, pages 22466--22474.

\bibitem[{Yannakoudakis and Briscoe(2012)}]{yannakoudakis2012lenthbased}
Helen Yannakoudakis and Ted Briscoe. 2012.
\newblock Modeling coherence in {ESOL} learner texts.
\newblock In \emph{Proceedings of the Seventh Workshop on Building Educational Applications Using {NLP}}, pages 33--43.

\bibitem[{Yannakoudakis et~al.(2011)Yannakoudakis, Briscoe, and Medlock}]{yannakoudakis2011CLCFCE}
Helen Yannakoudakis, Ted Briscoe, and Ben Medlock. 2011.
\newblock A new dataset and method for automatically grading {ESOL} texts.
\newblock In \emph{Proceedings of the 49th Annual Meeting of the Association for Computational Linguistics: Human Language Technologies}, pages 180--189.

\bibitem[{Ye et~al.(2025)Ye, Wang, Zou, Yan, Wang, Zheng, Xu, King, Yu, and Wen}]{ye2025position}
Jingheng Ye, Shen Wang, Deqing Zou, Yibo Yan, Kun Wang, Hai-Tao Zheng, Zenglin Xu, Irwin King, Philip~S Yu, and Qingsong Wen. 2025.
\newblock Position: Llms can be good tutors in foreign language education.
\newblock \emph{arXiv preprint arXiv:2502.05467}.

\bibitem[{Young et~al.(2024)Young, Chen, Li, Huang, Zhang, Zhang, Li, Zhu, Chen, Chang et~al.}]{young2024yi}
Alex Young, Bei Chen, Chao Li, Chengen Huang, Ge~Zhang, Guanwei Zhang, Heng Li, Jiangcheng Zhu, Jianqun Chen, Jing Chang, and 1 others. 2024.
\newblock Yi: Open foundation models by 01. ai.
\newblock \emph{arXiv preprint arXiv:2403.04652}.

\bibitem[{Yuan et~al.(2025)Yuan, Li, and Zhao}]{yuan2025survey}
Yuan Yuan, Zhaojian Li, and Bin Zhao. 2025.
\newblock A survey of multimodal learning: Methods, applications, and future.
\newblock \emph{ACM Computing Surveys}.

\bibitem[{Yuan et~al.(2024)Yuan, Chen, Wang, Yu, Peng, and Lou}]{yuan2024transagent}
Zhiqiang Yuan, Weitong Chen, Hanlin Wang, Kai Yu, Xin Peng, and Yiling Lou. 2024.
\newblock Transagent: An llm-based multi-agent system for code translation.
\newblock \emph{arXiv preprint arXiv:2409.19894}.

\bibitem[{Zhang et~al.(2025)Zhang, Chen, Wan, Chang, Cheng, Wang, Hu, and Bai}]{zhang2025evoflow}
Guibin Zhang, Kaijie Chen, Guancheng Wan, Heng Chang, Hong Cheng, Kun Wang, Shuyue Hu, and Lei Bai. 2025.
\newblock Evoflow: Evolving diverse agentic workflows on the fly.
\newblock \emph{arXiv preprint arXiv:2502.07373}.

\bibitem[{Zhang et~al.(2024{\natexlab{a}})Zhang, Yue, Li, Yun, Wan, Wang, Cheng, Yu, and Chen}]{zhang2024agentprune}
Guibin Zhang, Yanwei Yue, Zhixun Li, Sukwon Yun, Guancheng Wan, Kun Wang, Dawei Cheng, Jeffrey~Xu Yu, and Tianlong Chen. 2024{\natexlab{a}}.
\newblock Cut the crap: An economical communication pipeline for llm-based multi-agent systems.
\newblock \emph{arXiv preprint arXiv:2410.02506}.

\bibitem[{Zhang et~al.(2024{\natexlab{b}})Zhang, Yue, Sun, Wan, Yu, Fang, Wang, Chen, and Cheng}]{zhang2024g-designer}
Guibin Zhang, Yanwei Yue, Xiangguo Sun, Guancheng Wan, Miao Yu, Junfeng Fang, Kun Wang, Tianlong Chen, and Dawei Cheng. 2024{\natexlab{b}}.
\newblock G-designer: Architecting multi-agent communication topologies via graph neural networks.
\newblock \emph{arXiv preprint arXiv:2410.11782}.

\bibitem[{Zheng et~al.(2024)Zheng, Chen, Yan, Zou, and Hu}]{zheng2024reefknot}
Kening Zheng, Junkai Chen, Yibo Yan, Xin Zou, and Xuming Hu. 2024.
\newblock Reefknot: A comprehensive benchmark for relation hallucination evaluation, analysis and mitigation in multimodal large language models.
\newblock \emph{arXiv preprint arXiv:2408.09429}.

\bibitem[{Zhou et~al.(2024)Zhou, Yan, Zou, Wang, Liu, and Hu}]{zhou2024mitigating}
Guanyu Zhou, Yibo Yan, Xin Zou, Kun Wang, Aiwei Liu, and Xuming Hu. 2024.
\newblock Mitigating modality prior-induced hallucinations in multimodal large language models via deciphering attention causality.
\newblock \emph{arXiv preprint arXiv:2410.04780}.

\bibitem[{Zou et~al.(2024)Zou, Wang, Yan, Huang, Zheng, Chen, Tang, and Hu}]{zou2024look}
Xin Zou, Yizhou Wang, Yibo Yan, Sirui Huang, Kening Zheng, Junkai Chen, Chang Tang, and Xuming Hu. 2024.
\newblock Look twice before you answer: Memory-space visual retracing for hallucination mitigation in multimodal large language models.
\newblock \emph{arXiv preprint arXiv:2410.03577}.

\bibitem[{Zou et~al.(2025)Zou, Yan, Hao, Hu, Wen, Liu, Zhang, Li, Li, Zheng et~al.}]{zou2025deep}
Xingchen Zou, Yibo Yan, Xixuan Hao, Yuehong Hu, Haomin Wen, Erdong Liu, Junbo Zhang, Yong Li, Tianrui Li, Yu~Zheng, and 1 others. 2025.
\newblock Deep learning for cross-domain data fusion in urban computing: Taxonomy, advances, and outlook.
\newblock \emph{Information Fusion}, 113:102606.

\end{thebibliography}
\clearpage
\appendix

\section{More Related Work}
\label{sec:more_related_work}

\subsection{AES Datasets}
\label{sec:dataset}
Table \ref{tab:AES datasets} summarizes widely used AES datasets in terms of dataset size, number of essay topics, modality, and trait-level annotations. Most existing datasets (\textit{e.g.,} ASAP, CLC-FCE, TOEFL11) are unimodal and offer either holistic scores or a limited number of traits, primarily focusing on text-based prompts.
Recently, EssayJudge \cite{su2025essayjudge} has been introduced as a multimodal benchmark that includes both textual and visual inputs, covering 125 topics and annotated across ten fine-grained scoring traits. This enables more comprehensive evaluation of AES systems, especially those leveraging MLLMs.

\subsection{Multimodal Large Language Models}
\label{app:MLLM}
MLLMs have experienced rapid development in recent years and have been widely adopted across various domains \cite{yuan2025survey,song2025bridge,yan2024survey}. Their core advantage lies in the ability to jointly process visual and textual inputs to handle a range of complex tasks \cite{xi2023rise, huo2024mmneuron, yan2024urbanclip, yan2024georeasoner, zou2025deep, dang2024explainable,huo2025mmunlearner,chen2025safeeraser}. On the proprietary side, MLLMs such as GPT-4o \cite{openai2024gpt4ocard} and Gemini-1.5 \cite{gemini} have demonstrated state-of-the-art performance in multimodal reasoning, instruction following, and question answering tasks \cite{chang2024survey, yan2024errorradar, yan2024survey, zheng2024reefknot, yan2025position}. Meanwhile, open-source MLLMs have made notable advances in terms of accessibility and modularity. LLaVA-NEXT \cite{liu2024llavanext} employs pretrained encoders and adapters to align vision and language representations efficiently. Other representative MLLMs—such as Qwen2.5-VL \cite{qwen2_52025expanding}, DeepSeek-VL \cite{lu2024deepseek}, InternVL \cite{internvl2.5}, Yi-VL \cite{young2024yi}, LLaMA3-VL \cite{grattafiori2024llama3herdmodels}, and MiniCPM-V \cite{hu2024minicpmunveilingpotentialsmall}—have introduced a variety of fusion mechanisms, including visual projection heads, mixture-of-experts architectures, and image-grounded token masking. These MLLMs have been applied to a wide range of domains, including education and medical diagnostics \cite{qu2025tool, zou2024look, zhou2024mitigating, huang2024miner}, showcasing the expanding scope and depth of MLLM capabilities. Building further on the powerful capabilities of MLLM, the community is also focusing on exploring multi-agent scenarios \cite{yan2025mathagent,zhang2024g-designer,zhang2025evoflow,zhang2024agentprune}.

Building upon these diverse MLLMs, our proposed \agent multi-agent framework flexibly incorporates different MLLMs as the backbone for each agent module, enabling collaborative interaction between student and teacher MLLMs to enhance the accuracy and robustness of AES.

\section{Additional Experimental Details}
\subsection{Trait-Specific Rubrics}
\label{app:rubrics}
In this section, we introduce the rubrics of the 10 traits which is similar to EssayJudge. The rubrics are detailed in Table \ref{tab:AC} to Table \ref{tab:PA}. Each trait is assessed using a numerical score ranging from 0 to 5. A score of 5 represents high-quality performance with respect to the trait being evaluated, while a score of 0 represents low-quality performance in the same regard.

\subsection{Prompt for \agent framework}
\label{app:prompt}
Our agent-based \agent framework consists of three modules—Initial Scorer, Feedback Pool Generator, and Reflective Scorer—each employing customized prompt designs for their functional roles. While all agents operate under a unified trait-based rubric schema, the input structure and expected output vary to support multi-stage evaluation. The details are shown in Figure \ref{fig:initial scorer prompt example} to \ref{fig:reflective scorer prompt example}.

\subsection{Model Sources}
\label{app:sources}
Table \ref{tab:mllm hyperparams} details specific sources for the various student MLLMs. The hyperparameters for the experiments are set to their default values unless specified otherwise.

\subsection{Average Trait-Specific Score Comparison}
\label{app:average score comparison}
\begin{figure*}[htb!]
  \centering
  \includegraphics[width=1\textwidth]{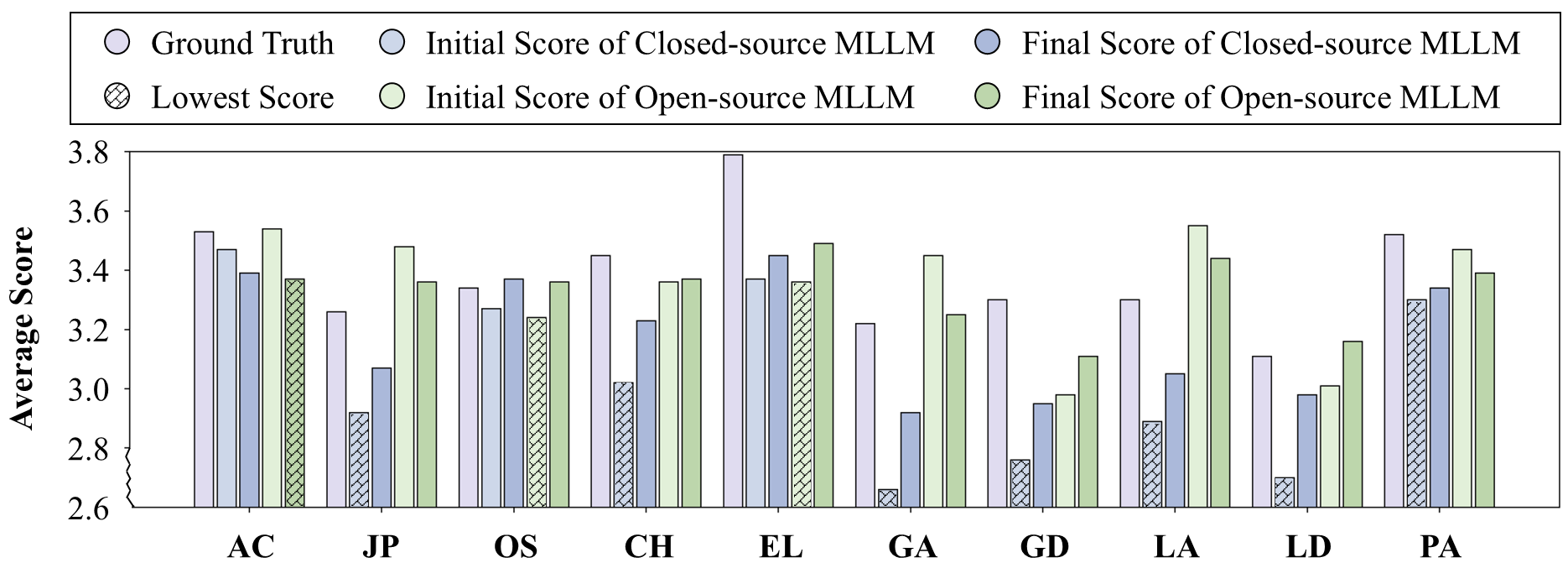}
  \caption{Average trait-specific scores assigned by closed-source and open-source MLLMs at both the initial stage and after revision through the \agent framework.}
  \label{fig:average comparison}
  \vspace{-4mm}
\end{figure*}

\begin{figure*}[htb!]
  \centering
  \includegraphics[width=0.5\textwidth]{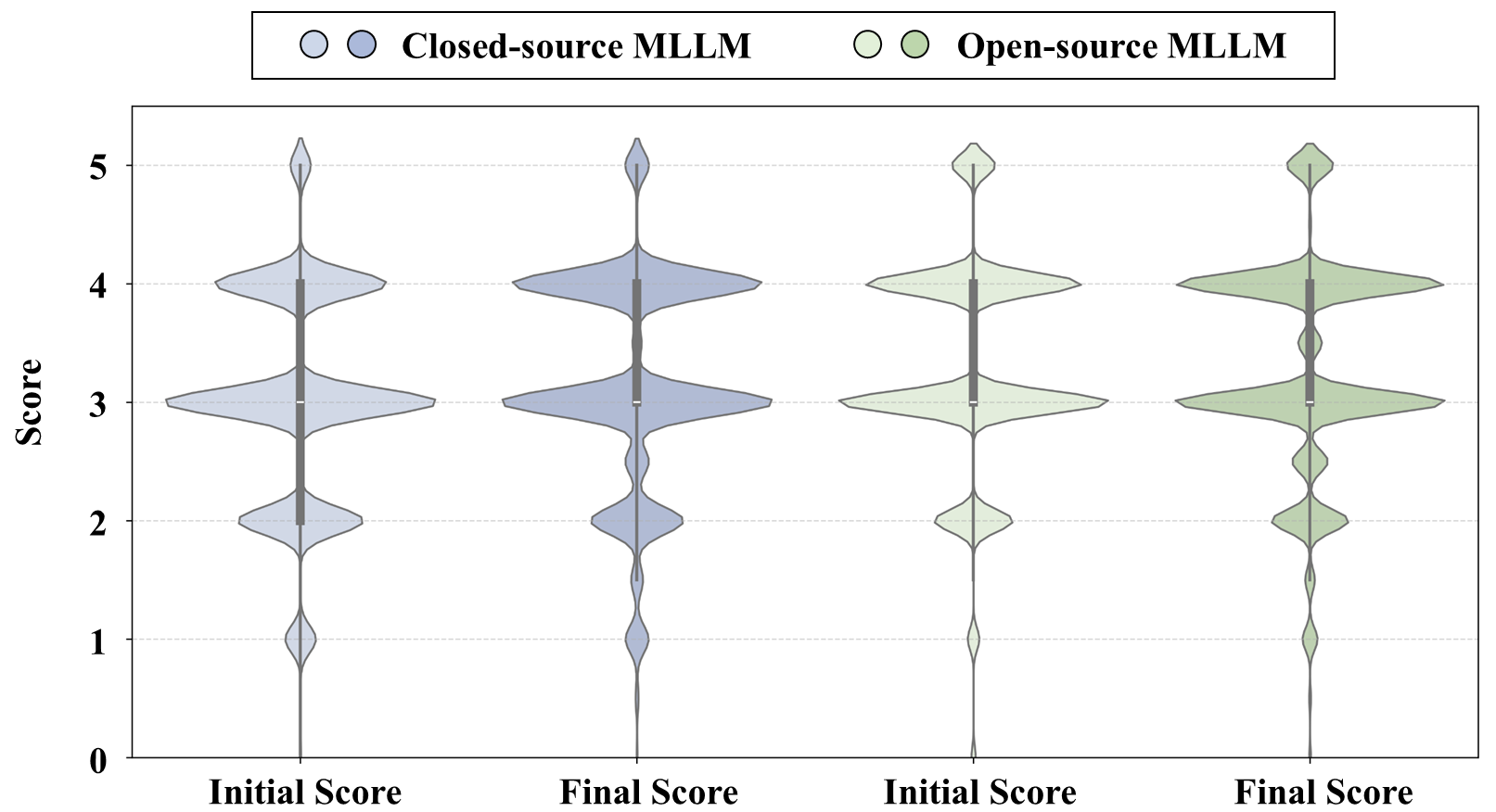}
  \caption{Score distributions of closed-source and open-source MLLMs at both the initial scoring stage and after revision through the \agent framework.}
  \label{fig:violin_all_models_distributon}
  \vspace{-4mm}
\end{figure*}

Closed-source MLLMs tend to adopt a more rigorous scoring strategy compared to open-source MLLMs. This trend is supported by both quantitative and distributional evidence. First, as shown in the Figure \ref{fig:average comparison}, closed-source models consistently assign lower average scores than open-source models across most traits, regardless of whether the scores are initial or final \cite{su2025essayjudge, kundu2024largelanguagemodelsgood}. Second, Figure \ref{fig:violin_all_models_distributon} reveals that closed-source models exhibit slightly higher score variance (0.81 vs. 0.79), indicating a broader and possibly more cautious distribution of judgments. Together, these findings suggest that closed-source MLLMs are more aligned with rigorous rubric interpretation, both when directly scoring and when acting as student models within the \agent framework.\subsection{More examples of Findings}
\label{app:more_examples_comparison}
\begin{figure*}[t]
    \centering
    \begin{minipage}{0.48\linewidth}
        \centering
        \includegraphics[width=\linewidth]{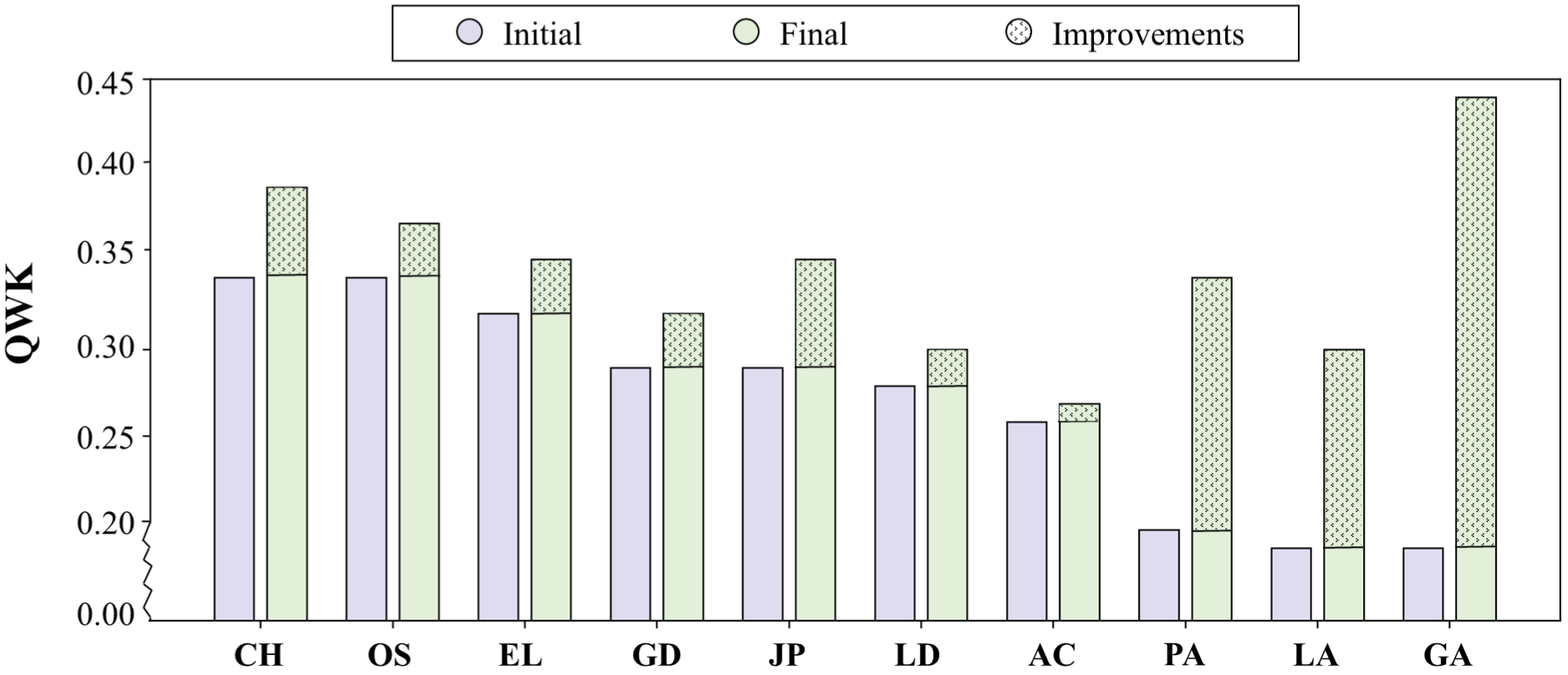}
        \caption*{(a) Qwen2.5-VL-3B }
    \end{minipage}
    \hfill
    \begin{minipage}{0.48\linewidth}
        \centering
        \includegraphics[width=\linewidth]{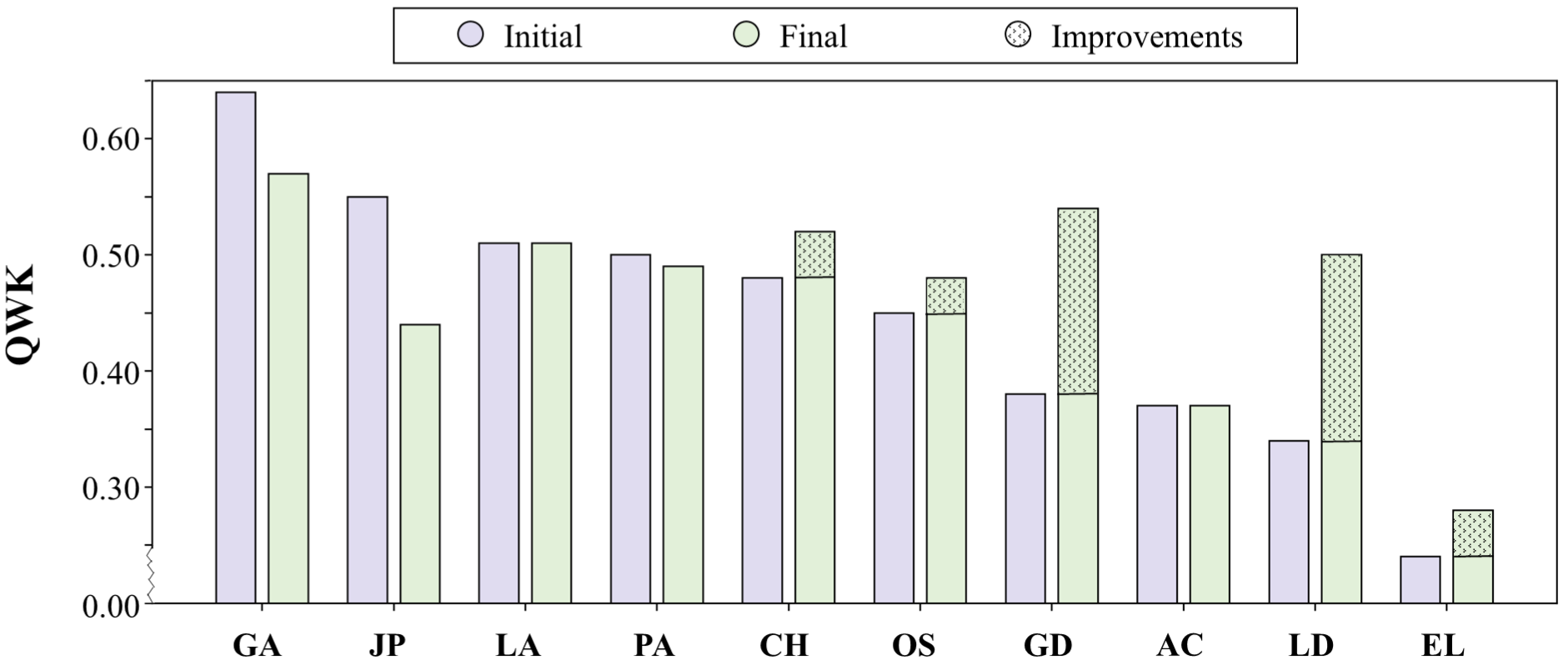}
        \caption*{(b) GPT-4o-mini}
    \end{minipage}
    \caption{Improvements of QWK score across all traits based on different student MLLM.}
    \label{fig:more_case_comparison}
\end{figure*}

Beyond the examples of Claude-3.5-Sonnet and InternVL2.5-8B presented in the main paper, additional cases support this observation, which is shown in Figure \ref{fig:more_case_comparison}.

\section{More Essay Scoring examples}
\label{app:case_study}
\begin{figure*}[htb!]
  \centering
  \includegraphics[width=1\textwidth]{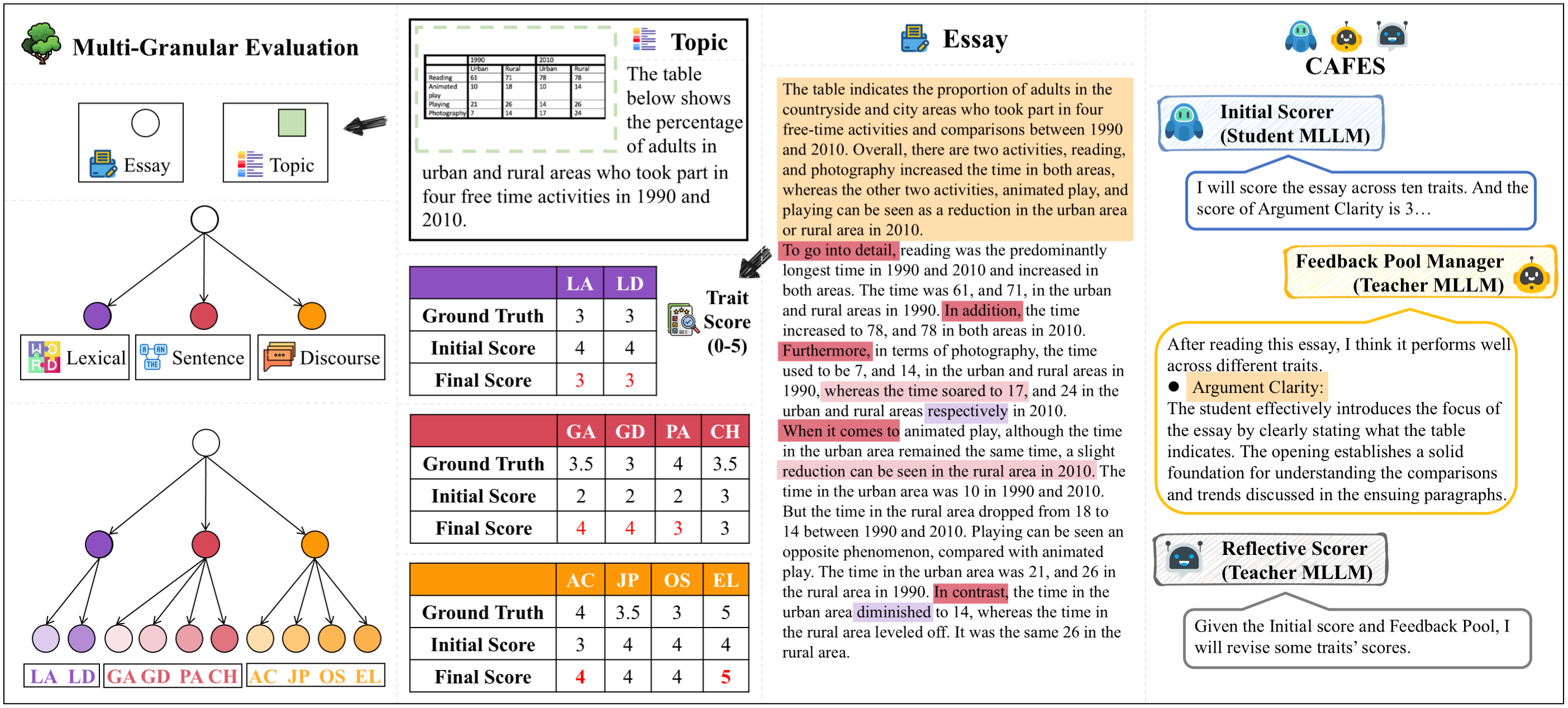}
  \caption{A case study illustrating \agent’s score revision process. And the student MLLM is Claude-3.5-Sonnet, and the teacher MLLM is GPT-4o.}
  \label{fig:case_study_gemini}
  \vspace{-4mm}
\end{figure*}

\begin{figure*}[htb!]
  \centering
  \includegraphics[width=1\textwidth]{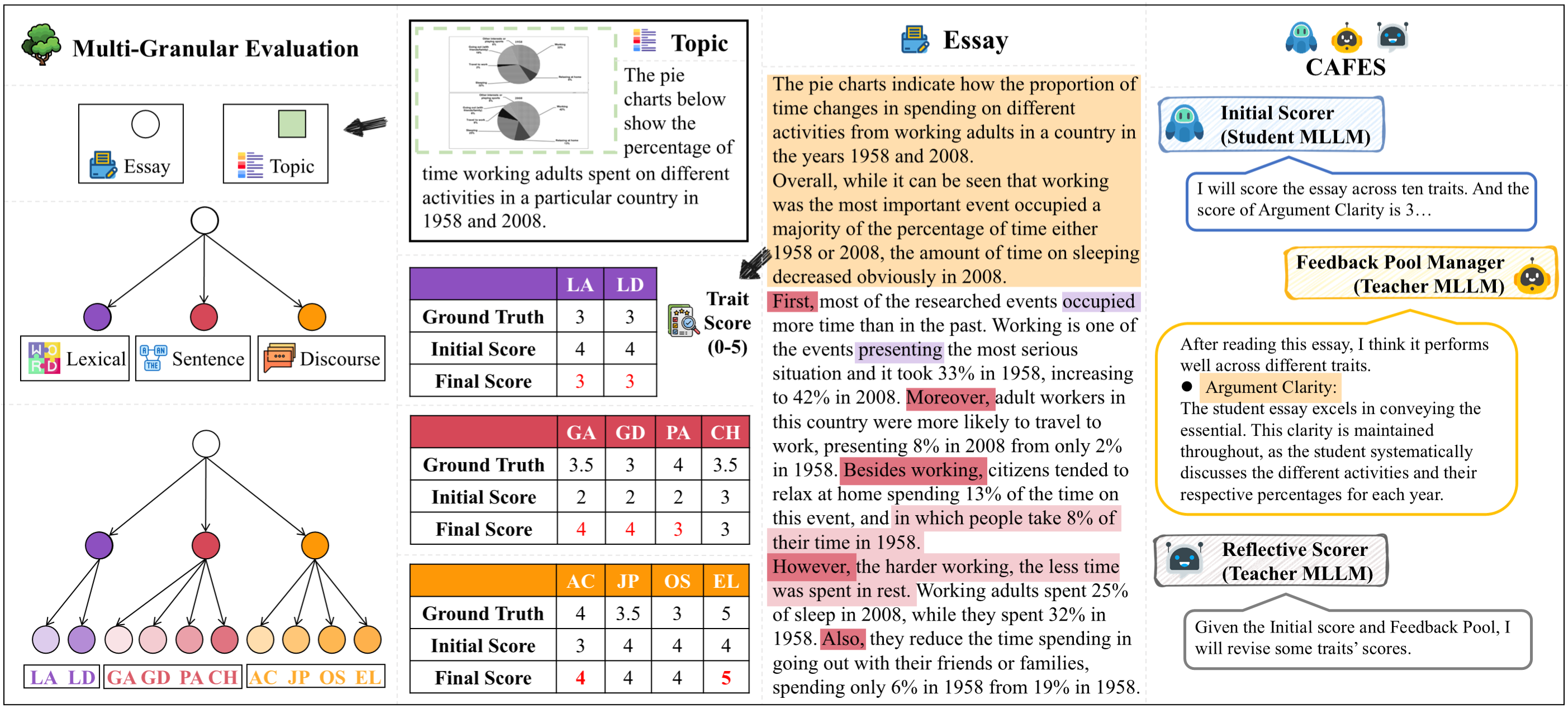}
  \caption{A case study illustrating \agent’s score revision process. And the student MLLM is GPT-4o-mini, and the teacher MLLM is GPT-4o.}
  \label{fig:case_study_gpt4omini}
  \vspace{-4mm}
\end{figure*}

\begin{figure*}[htb!]
  \centering
  \includegraphics[width=1\textwidth]{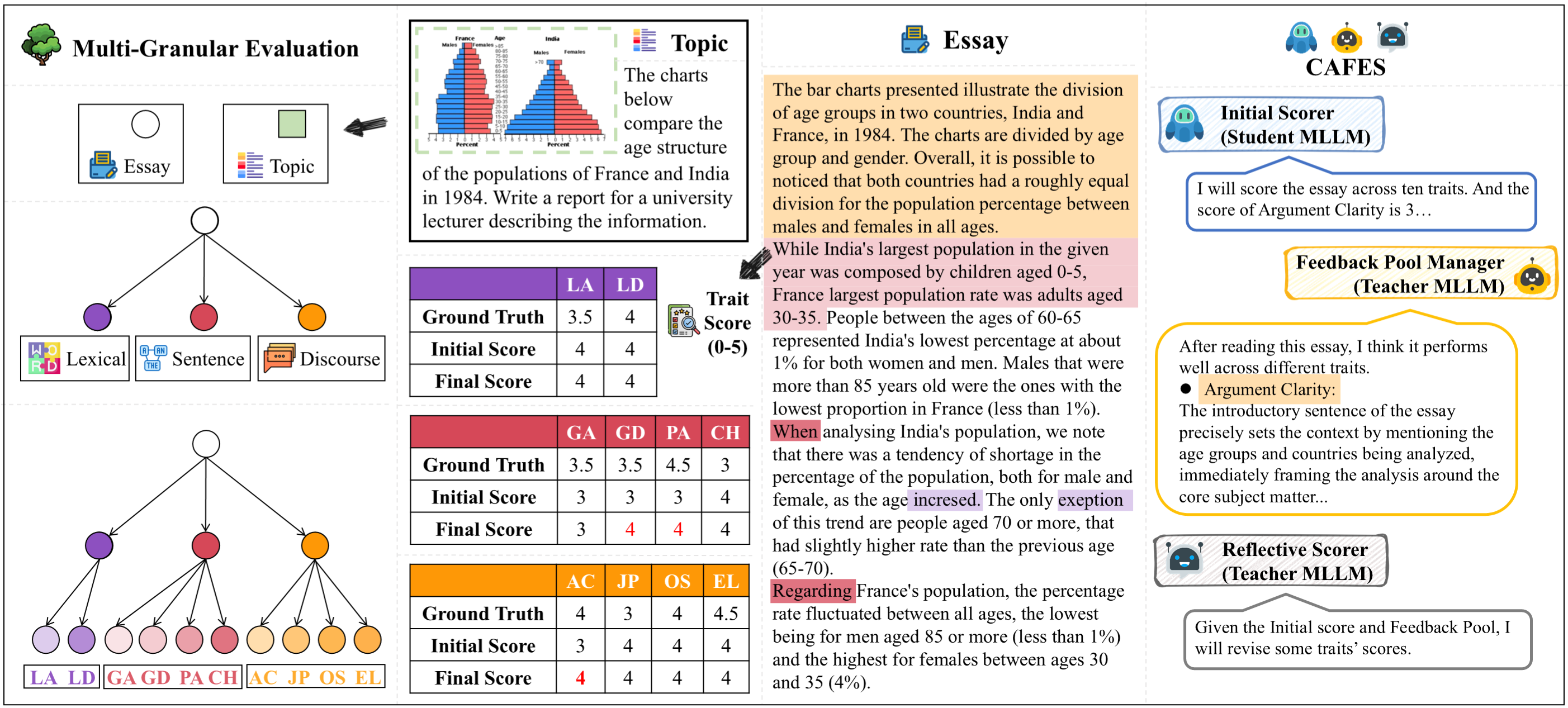}
  \caption{A case study illustrating \agent’s score revision process. And the student MLLM is Qwen2.5-VL-32B, and the teacher MLLM is GPT-4o.}
  \label{fig:case_study_qwen32b}
  \vspace{-4mm}
\end{figure*}

To further illustrate the effectiveness of our multi-agent AES framework, we include several additional essay cases (as shown in Figure \ref{fig:case_study_gemini} to \ref{fig:case_study_qwen32b}). Each example consists of the essay topic, student's essays, initial scores, feedback pool, and final revised scores after reflection. These examples highlight different error types, model reasoning behaviors, and improvement patterns across traits.

\begin{figure}[htbp]
\centering
\begin{minipage}{\columnwidth} 
    \centering 
    \small
    \renewcommand\tabcolsep{2pt} 
    \renewcommand\arraystretch{1.1}
    \resizebox{\columnwidth}{!}{
        \begin{tabular}{lccccc}
            \toprule
            \textbf{Benchmarks} & \textbf{Venue} & \textbf{Size} & \textbf{\#Topics} & \textbf{Modality} & \textbf{\#Traits} \\
            \midrule
            $\text{ASAP}_{\text{AES}}$ \cite{cozma2018ASAP} & ACL & 17,450 & 8 & T & 0 \\
            ASAP++ \cite{mathias2018asap++} & ACL & 10,696 & 6 & T & 8 \\
            CLC-FCE \cite{yannakoudakis2011CLCFCE} & ACL & 1,244 & 10 & T & 0 \\
            TOEFL11 \cite{lee2024TOEFL11} & EMNLP & 1,100 & 8 & T & 0 \\
            ICLE \cite{granger2009icle} & COLING & 3,663 & 48 & T & 4\\
            AAE \cite{stab2014AAE} & COLING & 102 & 101 & T & 1 \\
            ICLE++ \cite{li2024icle++} & NAACL & 1,008 & 10 & T & 10 \\
            CREE \cite{bailey2008CREE} & BEA & 566 & 75 & T & 1 \\
            \midrule
            EssayJudge \cite{su2025essayjudge} & - & 1054 & 125 & T, I & 10 \\
            \bottomrule
        \end{tabular}
    }
    \captionof{table}{Comparison between previous AES datasets. }
    \label{tab:AES datasets}
    \vspace{-2mm}
\end{minipage}
\end{figure}

\begin{table*}[htb!]
  \centering
  \begin{tabularx}{\textwidth}{>{\centering\arraybackslash}p{0.7cm}>{\raggedright\arraybackslash}X} 
    \toprule
    \multicolumn{1}{c}{\textbf{Score}} & \multicolumn{1}{c}{\textbf{Scoring Criteria}} \\ 
    \midrule
    5 & The central argument is clear, and the first paragraph clearly outlines the topic of the image and question, providing guidance with no ambiguity.   \\ 
    \midrule    
    4 & The central argument is clear, and the first paragraph mentions the topic of the image and question, but the guidance is slightly lacking or the expression is somewhat vague.  \\ 
    \midrule    
    3 & The argument is generally clear, but the expression is vague, and it doesn't adequately guide the rest of the essay. \\ 
    \midrule    
    2 & The argument is unclear, the description is vague or incomplete, and it doesn't guide the essay.   \\ 
    \midrule    
    1 & The argument is vague, and the first paragraph fails to effectively summarize the topic of the image or question.    \\ 
    \midrule    
    0 & No central argument is presented, or the essay completely deviates from the topic and image.    \\ 
    \bottomrule
  \end{tabularx}
  \caption{Rubrics for evaluating the argument clarity of the essays.}
  \label{tab:AC}
\end{table*}

\begin{table*}[htb!]
  \centering
  \begin{tabularx}{\textwidth}{>{\centering\arraybackslash}p{0.7cm}>{\raggedright\arraybackslash}X} 
    \toprule
    \multicolumn{1}{c}{\textbf{Score}} & \multicolumn{1}{c}{\textbf{Scoring Criteria}} \\ 
    \midrule
    5  & Transitions between sentences are natural, and logical connections flow smoothly; appropriate use of linking words and transitional phrases.   \\ 
    \midrule
    4  & Sentences are generally coherent, with some transitions slightly awkward; linking words are used sparingly but are generally appropriate.   \\ 
    \midrule    
    3  & The logical connection between sentences is not smooth, with some sentences jumping or lacking flow; linking words are used insufficiently or inappropriately. \\ 
    \midrule    
    2  & Logical connections are weak, sentence connections are awkward, and linking words are either used too little or excessively.    \\ 
    \midrule    
    1  & There is almost no logical connection between sentences, transitions are unnatural, and linking words are very limited or incorrect.     \\ 
    \midrule    
    0  & No coherence at all, with logical confusion between sentences.    \\ 
    \bottomrule
  \end{tabularx}
  \caption{Rubrics for evaluating the coherence of the essays.}
  \label{tab:CH}
\end{table*}

\begin{table*}[htbp]
  \centering
  \begin{tabularx}{\textwidth}{>{\centering\arraybackslash}p{0.7cm}>{\raggedright\arraybackslash}X} 
    \toprule
    \multicolumn{1}{c}{\textbf{Score}} & \multicolumn{1}{c}{\textbf{Scoring Criteria}} \\ 
    \midrule
    5 & Word count is 150 words or more, with the content being substantial and without obvious excess or brevity.    \\ 
    \midrule
    4 & Word count is around 150 words, but slightly off (within 10 words), and the content is complete.    \\ 
    \midrule
    3 & Word count is noticeably too short or too long, and the content is not sufficiently substantial or is somewhat lengthy.  \\ 
    \midrule
    2 & Word count deviates significantly, failing to fully cover the requirements of the prompt.    \\ 
    \midrule
    1 & Word count is far below the requirement, and the content is incomplete.     \\ 
    \midrule
    0 & Word count is severely insufficient or excessive, making it impossible to meet the requirements of the prompt.    \\ 
    \bottomrule
  \end{tabularx}
  \caption{Rubrics for evaluating the essay length of the essays.}
  \label{tab:EL}
\end{table*}

\begin{table*}[htbp]
  \centering
  \begin{tabularx}{\textwidth}{>{\centering\arraybackslash}p{0.7cm}>{\raggedright\arraybackslash}X} 
    \toprule
    \multicolumn{1}{c}{\textbf{Score}} & \multicolumn{1}{c}{\textbf{Scoring Criteria}} \\ 
    \midrule
    5 & Sentence structure is accurate with no grammatical errors; both simple and complex sentences are error-free.    \\    
    \midrule
    4 & Sentence structure is generally accurate, with occasional minor errors that do not affect understanding; some errors in complex sentence structures.   \\    
    \midrule
    3 & Few grammatical errors, but more noticeable errors that affect understanding; simple sentences are accurate, but complex sentences frequently contain errors.  \\   
    \midrule
    2 & Numerous grammatical errors, with sentence structure affecting understanding; simple sentences are occasionally correct, but complex sentences have frequent errors.    \\  
    \midrule
    1 & A large number of grammatical errors, with sentence structure severely affecting understanding; sentence structure is unstable, and even simple sentences contain mistakes.     \\  
    \midrule
    0 & Sentence structure is completely incorrect, nonsensical, and difficult to understand.    \\ 
    \bottomrule
  \end{tabularx}
  \caption{Rubrics for evaluating the grammatical accuracy of the essays.}
  \label{tab:GA}
\end{table*}

\begin{table*}[htbp]
  \centering
  \begin{tabularx}{\textwidth}{>{\centering\arraybackslash}p{0.7cm}>{\raggedright\arraybackslash}X} 
    \toprule
    \multicolumn{1}{c}{\textbf{Score}} & \multicolumn{1}{c}{\textbf{Scoring Criteria}} \\ 
    \midrule
    5  & Uses a variety of sentence structures, including both simple and complex sentences, with flexible use of clauses and compound sentences, demonstrating rich sentence variation.     \\   
    \midrule
    4  & Generally uses a variety of sentence structures, with appropriate use of common clauses and compound sentences. Sentence structures vary, though some sentence types lack flexibility.    \\  
    \midrule
    3  & Uses a variety of sentence structures, but with limited use of complex sentences, which often contain errors. Sentence variation is somewhat restricted.  \\     
    \midrule
    2  & Sentence structures are simple, primarily relying on simple sentences, with occasional attempts at complex sentences, though errors occur frequently.    \\     
    \midrule
    1  & Sentence structures are very basic, with almost no complex sentences, and even simple sentences contain errors.     \\  
    \midrule
    0  & Only uses simple, repetitive sentences with no complex sentences, resulting in rigid sentence structures.    \\ 
    \bottomrule
  \end{tabularx}
  \caption{Rubrics for evaluating the grammatical diversity of the essays.}
  \label{tab:GD}
\end{table*}

\begin{table*}[htbp]
  \centering
  \begin{tabularx}{\textwidth}{>{\centering\arraybackslash}p{0.7cm}>{\raggedright\arraybackslash}X} 
    \toprule
    \multicolumn{1}{c}{\textbf{Score}} & \multicolumn{1}{c}{\textbf{Scoring Criteria}} \\ 
    \midrule
    5  & Fully addresses and accurately analyzes all important information in the image and prompt (\textit{e.g.}, data turning points, trends); argumentation is in-depth and logically sound.      \\    
    \midrule
    4  & Addresses most of the important information in the image and prompt, with reasonable analysis but slight shortcomings; argumentation is generally logical.     \\     
    \midrule
    3  &Addresses some important information in the image and prompt, but analysis is insufficient; argumentation is somewhat weak. \\     
    \midrule
    2  & Mentions a small amount of important information in the image and prompt, with simple or incorrect analysis; there are significant logical issues in the argumentation.   \\  
    \midrule
    1  & Only briefly mentions important information in the image and prompt or makes clear analytical errors, lacking reasonable reasoning.      \\   
    \midrule
    0  & Fails to mention key information from the image and prompt, lacks any argumentation, and is logically incoherent.   \\ 
    \bottomrule
  \end{tabularx}
  \caption{Rubrics for evaluating the justifying persuasiveness of the essays.}
  \label{tab:JP}
\end{table*}

\begin{table*}[htbp]
  \centering
  \begin{tabularx}{\textwidth}{>{\centering\arraybackslash}p{0.7cm}>{\raggedright\arraybackslash}X} 
    \toprule
    \multicolumn{1}{c}{\textbf{Score}} & \multicolumn{1}{c}{\textbf{Scoring Criteria}} \\ 
    \midrule
    5 & Vocabulary is accurately chosen, with correct meanings and spelling, and minimal errors; words are used precisely to convey the intended meaning.     \\ 
    \midrule
    4 & Vocabulary is generally accurate, with occasional slight meaning errors or minor spelling mistakes, but they do not affect overall understanding; words are fairly precise.     \\ 
    \midrule
    3 & Vocabulary is mostly correct, but frequent minor errors or spelling mistakes affect some expressions; word choice is not fully precise.  \\ 
    \midrule
    2 & Vocabulary is inaccurate, with significant meaning errors and frequent spelling mistakes, affecting understanding.    \\ 
    \midrule
    1 & Vocabulary is severely incorrect, with unclear meanings and noticeable spelling errors, making comprehension difficult.       \\ 
    \midrule
    0 & Vocabulary choice and spelling are completely incorrect, and the intended meaning is unclear or impossible to understand.   \\  
    \bottomrule
  \end{tabularx}
  \caption{Rubrics for evaluating the lexical accuracy of the essays.}
  \label{tab:LA}
\end{table*}

\begin{table*}[htbp]
  \centering
  \begin{tabularx}{\textwidth}{>{\centering\arraybackslash}p{0.7cm}>{\raggedright\arraybackslash}X} 
    \toprule
    \multicolumn{1}{c}{\textbf{Score}} & \multicolumn{1}{c}{\textbf{Scoring Criteria}} \\ 
    \midrule
    5  & Vocabulary is rich and diverse, with a wide range of words used flexibly, avoiding repetition.   \\ 
    \midrule
    4  & Vocabulary diversity is good, with a broad range of word choices, occasional repetition, but overall flexible expression.     \\ 
    \midrule
    3  &Vocabulary diversity is average, with some variety in word choice but limited, with frequent repetition.  \\ 
    \midrule
    2  & Vocabulary is fairly limited, with a lot of repetition and restricted word choice.   \\ 
    \midrule
    1  & Vocabulary is very limited, with frequent repetition and an extremely narrow range of words.       \\ 
    \midrule
    0  & Vocabulary is monotonous, with almost no variation, failing to demonstrate vocabulary diversity.  \\  
    \bottomrule
  \end{tabularx}
  \caption{Rubrics for evaluating the lexical diversity of the essays.}
  \label{tab:LD}
\end{table*}

\begin{table*}[htbp]
  \centering
  \begin{tabularx}{\textwidth}{>{\centering\arraybackslash}p{0.7cm}>{\raggedright\arraybackslash}X} 
    \toprule
    \multicolumn{1}{c}{\textbf{Score}} & \multicolumn{1}{c}{\textbf{Scoring Criteria}} \\ 
    \midrule
    5 & The essay has a well-organized structure, with clear paragraph divisions, each focused on a single theme. There are clear topic sentences and concluding sentences, and transitions between paragraphs are natural.  \\ 
    \midrule
    4 & The structure is generally reasonable, with fairly clear paragraph divisions, though transitions may be somewhat awkward and some paragraphs may lack clear topic sentences.      \\ 
    \midrule
    3 &The structure is somewhat disorganized, with unclear paragraph divisions, a lack of topic sentences, or weak logical flow.  \\ 
    \midrule
    2 & The structure is unclear, with improper paragraph divisions and poor logical coherence.    \\ 
    \midrule
    1 & The paragraph structure is chaotic, with most paragraphs lacking clear topic sentences and disorganized content.       \\ 
    \midrule
    0 & No paragraph structure, content is jumbled, and there is a complete lack of logical connections. \\
    \bottomrule
  \end{tabularx}
  \caption{Rubrics for evaluating the organizational structure of the essays.}
  \label{tab:OS}
\end{table*}

 \begin{table*}[htbp]
  \centering
  \begin{tabularx}{\textwidth}{>{\centering\arraybackslash}p{0.7cm}>{\raggedright\arraybackslash}X} 
    \toprule
    \multicolumn{1}{c}{\textbf{Score}} & \multicolumn{1}{c}{\textbf{Scoring Criteria}} \\ 
    \midrule
    5 & Punctuation is used correctly throughout, adhering to standard rules with no errors. \\ 
    \midrule
    4 & Punctuation is mostly correct, with occasional minor errors that do not affect understanding.     \\ 
    \midrule
    3 &Punctuation is generally correct, but there are some noticeable errors that slightly affect understanding.  \\ 
    \midrule
    2 & There are frequent punctuation errors, some of which affect understanding.    \\ 
    \midrule
    1 & Punctuation errors are severe, significantly affecting comprehension.       \\ 
    \midrule
    0 & Punctuation is completely incorrect or barely used, severely hindering understanding. \\ 
    \bottomrule
  \end{tabularx}
  \caption{Rubrics for evaluating the punctuation accuracy of the essays.}
  \label{tab:PA}
\end{table*}

\begin{figure*}[th!]
    \centering
    \begin{tcolorbox}[colback=white, colframe=black, enhanced jigsaw, listing only, listing options={basicstyle=\rmfamily}]
        \textbf{Task Definition:} You are an experienced English writing examiner. Please evaluate the student's essay by assigning a score (0-5) for each of the ten traits and a confidence level (1–10) that reflects how certain you are about each score, where 1 is least certain and 10 is completely certain.
        \\[1em]
        \textbf{Rubrics:} \{Trait-specific corresponding rubrics\}
        \\[1em]
        \textbf{Below is the reference content:} \\
        Image: "\{image\}"\\
        Essay Topic: "\{question\}"\\
        Student's Essay: "\{essay\}" \\[1em]
        \textbf{Instruction:} Please provide your answer in the same style and format as the example. Use the exact trait names as shown (with proper capitalization)
        Return your response strictly in JSON format without any additional text, explanations, or code block delimiters (no triple backticks). 
    \end{tcolorbox}
    \caption{Prompt for Initial Scorer.}
    \label{fig:initial scorer prompt example}
\end{figure*}

\begin{figure*}[th!]
    \centering
    \begin{tcolorbox}[colback=white, colframe=black, enhanced jigsaw, listing only, listing options={basicstyle=\rmfamily}]
        \textbf{Task Definition:} You are an experienced English writing examiner. Your task is to provide detailed positive feedback on a student essay across ten traits: 
        Argument Clarity, Justifying Persuasiveness, Organizational Structure, Coherence, Essay Length, Grammatical Accuracy, Grammatical Diversity, Lexical Accuracy, Lexical Diversity, Punctuation Accuracy.
        \\[1em]
        \textbf{Rubrics:} \{Trait-specific corresponding rubrics\}
        \\[1em]
        \textbf{Below is the reference content:} \\
        Image: "\{image\}"\\
        Essay Topic: "\{question\}"\\
        Student's Essay: "\{essay\}" \\[1em]
        \textbf{Instruction:} please generate your feedback dimension by dimension.Your output must be in natural language paragraphs. Do not use JSON, code blocks, or bullet points. Start each dimension with the tag in square brackets, for example: [Argument Clarity]\\
        Sample Format:
        [Argument Clarity]
        The opening paragraph clearly introduces the topic of the image and outlines the overall trend, effectively setting up the structure for later analysis.
    \end{tcolorbox}
    \caption{Prompt for Feedback Pool Manager.}
    \label{fig:feedback manager prompt example}
\end{figure*}

\begin{figure*}[th!]
    \centering
    \begin{tcolorbox}[colback=white, colframe=black, enhanced jigsaw, listing only, listing options={basicstyle=\rmfamily}]
        \textbf{Task Definition:} You are evaluating a set of essay scores originally provided by another assistant reviewer. A detailed feedback report—including both positive and negative comments across 10 traits—is available for reference, but should not be treated as an absolute judgment. 
        Your task is to serve as a more careful and critical second-round reviewer. Do not assume the original scores are correct — examine each trait carefully and revise any score that appears inaccurate or unsupported by the essay.
        \\[1em]
        \textbf{Rubrics:} \{Trait-specific corresponding rubrics\}
        \\[1em]
        \textbf{Below is the reference content:} \\
        Image: "\{image\}"\\
        Essay Topic: "\{question\}"\\
        Student's Essay: "\{essay\}" \\
        Original Scores :"\{score\}"\\
        Feedback Report :"\{feedback\}"\\
        [1em]
        \textbf{Instruction:} If revision is needed, return only the affected dimensions with new scores and brief reasoning. Otherwise, confirm the original scores.
        Return your response strictly in JSON format without any additional text, explanations, or code block delimiters (no triple backticks). Only raw JSON is accepted.
        
    \end{tcolorbox}
    \caption{Prompt for Reflective Scorer.}
    \label{fig:reflective scorer prompt example}
\end{figure*}

\begin{table*}[htbp]
\small
\centering
\begin{tabular}{p{0.2\linewidth} | p{0.3\linewidth} | p{0.3\linewidth}}
\toprule
\textbf{MLLMs} & \textbf{Source} & \textbf{URL} \\
\midrule
InternVL2.5-2B & local checkpoint & \url{https://huggingface.co/OpenGVLab/InternVL2-2B} \\
\midrule
InternVL2.5-4B & local checkpoint & \url{https://huggingface.co/OpenGVLab/InternVL2-4B} \\
\midrule
InternVL2.5-8B & local checkpoint & \url{https://huggingface.co/OpenGVLab/InternVL2-8B} \\
\midrule
InternVL2.5-26B & local checkpoint &  \url{https://huggingface.co/OpenGVLab/InternVL2-26B}\\
\midrule
Qwen2.5-VL-3B  & local checkpoint & \url{https://huggingface.co/Qwen/Qwen2.5-VL-3B-Instruct} \\
\midrule
Qwen2.5-VL-32B  & local checkpoint & \url{https://huggingface.co/Qwen/Qwen2.5-VL-32B-Instruct} \\
\midrule
LLaMA-3.2-Vision-11B  & local checkpoint & \url{https://huggingface.co/meta-llama/Llama-3.2-11B-Vision-Instruct} \\
\midrule
LLaMA-3.2-Vision-90B & local checkpoint & \url{https://huggingface.co/meta-llama/Llama-3.2-90B-Vision-Instruct} \\
\midrule
Claude-3.5-Sonnet & \texttt{claude-3.5-sonnet-20241022} & \url{https://www.anthropic.com/claude/sonnet} \\
\midrule
Gemini-2.5-Flash  & \texttt{gemini-2.5-flash-preview-04-17} & \url{https://deepmind.google/technologies/gemini/flash} \\
\midrule
GPT-4o-mini & \texttt{gpt-4o-mini-2024-07-18} & \url{https://platform.openai.com/docs/models/gpt-4o-mini} \\
\bottomrule
\end{tabular}
\caption{Sources of our evaluated MLLMs.}
\label{tab:mllm hyperparams}
\end{table*}

\end{document}